\documentclass[10pt,journal,compsoc]{IEEEtran}
\ifCLASSOPTIONcompsoc
  \usepackage{cite}
  \usepackage{amsmath,amsfonts}
  \usepackage{enumitem}
  \usepackage{algorithm}
  \usepackage{algpseudocode}
  \usepackage{array}
  \usepackage{cite}
  \usepackage[caption=false,font=normalsize,labelfont=sf,textfont=sf]{subfig}
  \usepackage{textcomp}
  \usepackage{stfloats}       
  \usepackage{amssymb}        
  \usepackage{xcolor} 
  \usepackage{amsmath}        
  \usepackage{url}
  \usepackage{verbatim}
  \usepackage{graphicx}
  \usepackage{cite}
  \usepackage{multirow}
  \usepackage{booktabs}
  \usepackage{tabularx} 
  \usepackage[table]{xcolor}
  \usepackage[utf8]{inputenc}
\usepackage{listings} 
\usepackage{xcolor}   
\usepackage{caption}  
    \usepackage{booktabs} 
    
  \usepackage{hyperref}
  \hypersetup{
    colorlinks=true,       
    linkcolor=blue,        
    citecolor=blue,        
    urlcolor=blue,         
    pdfborder={0 0 0},     
}
\else
  \usepackage{cite}
\fi

\ifCLASSINFOpdf

\else

\fi

\begin{document}
\title{Learning to Use Imagination: Progress-Conditioned Future Utilization for World Action Models}
\author{Yijie~Zhu,
Zitong~Yu,~\IEEEmembership{Senior~Member,~IEEE,}
Wei~Li,
Hui~Ma,
Wen~Li,~\IEEEmembership{Member,~IEEE,}
Rui~Shao,~\IEEEmembership{Member,~IEEE,}
Liqiang~Nie,~\IEEEmembership{Senior~Member,~IEEE}
\IEEEcompsocitemizethanks{
\IEEEcompsocthanksitem Yijie Zhu, Zitong Yu, and Wei Li contributed equally to this work. Corresponding authors: Zitong Yu 
(email: yuzitong@gbu.edu.cn) and Rui Shao (email: shaorui@hit.edu.cn).
\IEEEcompsocthanksitem Yijie Zhu is with Harbin Institute of Technology (Shenzhen),
518055, China, and Great Bay University, Dongguan 523000, China. Zitong Yu and Hui Ma are with 
Great Bay University, Dongguan 523000, China. Wei Li, Rui Shao and Liqiang Nie are with Harbin Institute of Technology (Shenzhen), 518055, China. Wen Li is with the University of Electronic Science and Technology of China, Chengdu 611731, China.
}
}

\markboth{IEEE Transactions on Pattern Analysis and Machine Intelligence}
{Shell \MakeLowercase{\textit{et al.}}: Bare Demo of IEEEtran.cls for Computer Society Journals}
\IEEEtitleabstractindextext{%

\begin{abstract}
World Action Models (WAMs) extend Vision-Language-Action (VLA) models by
incorporating future visual dynamics into action generation. However, existing
WAMs often utilize imagined futures with limited adaptation to evolving
execution progress, potentially introducing distracting or unreliable
predictive cues. This limitation arises from two empirically identified forms of non-uniformity in future utility: \textbf{(i)} at the inter-progress level, the utility of imagined futures
varies across execution stages as control demands change; and \textbf{(ii)} at the
intra-progress level, individual future latents exhibit heterogeneous relevance
within the same progress state.
To address these limitations, we propose \textbf{ProWAM}, a
\textbf{Progress-Conditioned World Action Model} that introduces execution
progress as an explicit intermediate representation for adaptive imagination
utilization. ProWAM comprises two tightly coupled components:
\textbf{(1)} To obtain a reliable representation of execution progress, we propose the
\textbf{Self-Supervised Dual-Temporal Progress Encoder (SS-DTPE)}. SS-DTPE
couples short-term action--observation interaction modeling with long-term
recurrent progress aggregation to capture recent execution feedback and
accumulated task history. Rather than relying on explicit progress annotations,
SS-DTPE learns structured progress representations by exploiting the intrinsic
semantic and temporal structure of demonstrations through self-supervision.
\textbf{(2)} Conditioned on the progress representation from SS-DTPE, we propose the
\textbf{Hierarchical Progress-Conditioned Imagination Modulation (HPIM)} to
adapt imagination utilization to execution progress. HPIM operates at two
complementary levels: an inter-progress global modulation mechanism adapts
future utilization across execution stages, while an intra-progress relevance
mechanism differentiates individual future latents within each progress state.
Together, they enable stage-adaptive and latent-specific imagination utilization.
Extensive experiments on five simulation benchmarks and diverse real-world
tasks on two robotic platforms demonstrate consistent gains over strong VLA
and WAM baselines. Code and project page are available at:
\url{https://github.com/JiuTian-VL/ProWAM}.
\end{abstract}


}

\maketitle

\IEEEdisplaynontitleabstractindextext

\IEEEpeerreviewmaketitle
\vspace{-5pt}
\IEEEraisesectionheading{\section{Introduction}\label{sec:introduction}}
\label{sec:intro}

\begin{figure*}[t]
    \centering
    \includegraphics[width=\textwidth]{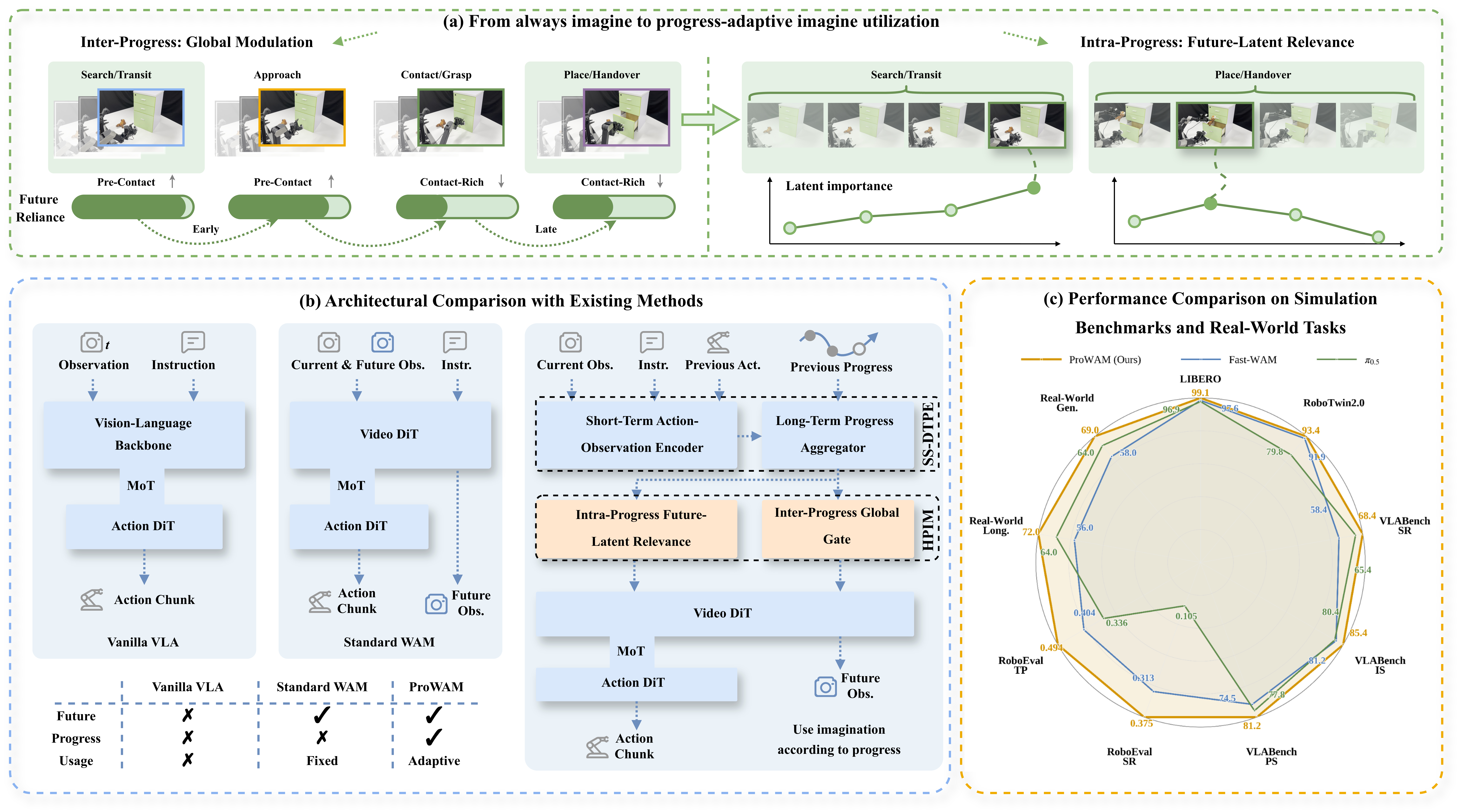}
    \vspace{-10pt}
   \caption{
\textbf{(a) Motivation:} Standard WAMs often adopt fixed imagination usage,
overlooking progress-dependent utility and unequal relevance among future
latents. ProWAM addresses these limitations through inter-progress global
modulation and intra-progress future-latent relevance.
\textbf{(b) Architecture:} Unlike Vanilla VLAs and Standard WAMs, ProWAM
explicitly models execution progress to adapt imagination utilization.
\textbf{(c) Performance:} ProWAM consistently outperforms representative
baselines across simulation and real-world manipulation tasks.}

    \label{fig:intro_overview}
    \vspace{-8pt}
\end{figure*}

\IEEEPARstart{V}{ision}-Language-Action (VLA) models have recently demonstrated
strong potential for generalizable robotic manipulation by learning unified
mappings from visual observations and language instructions to robot actions
\cite{kim2025fine,black2024pi_0,zhu2026h,li2026cogvla,li2026global,
shi2025memoryvla,song2025pd,zhang2026dreamvla,li2026semanticvla}.
Despite their impressive policy learning capabilities, conventional VLAs
primarily focus on action generation from the observed context, without
explicitly modeling how the visual scene may evolve during task execution.
This limitation has motivated growing interest in World Action Models (WAMs)
\cite{li2026causal,yuan2026fast,wang2026trust,huang2026noisegate,
li2026egowam,feng2026wam,li2026wam4d,jiang2026ckt,zhu2026dswam,wang2026dim},
which incorporate future visual dynamics into action generation within a unified
predictive framework. By jointly reasoning over anticipated visual states and
robot actions, WAMs provide predictive cues beyond the current observation,
supporting more informed decision making over extended manipulation horizons.

Despite these advances, existing WAMs have primarily focused on improving how
future information is modeled through richer representations, spatial grounding,
temporal memory, or more efficient inference
\cite{cen2025worldvla,bi2026motus,li2026egowam,li2026wam4d,
li2026light,yuan2026fast,wang2026dim,zhu2026dswam}.
While these efforts substantially advance future modeling, comparatively less
attention has been paid to how imagined futures should be adaptively utilized as
task execution evolves. In many existing WAMs, imagined futures are incorporated
through largely fixed or progress-agnostic interaction patterns, with limited
adaptation to the evolving execution progress. Such utilization may introduce
distracting or unreliable predictive cues when the relevance of future
information changes during task execution.

This motivates our key intuition that imagined futures
should not be utilized uniformly throughout robotic manipulation, but adapt to
the evolving execution progress. As conceptually illustrated in
Fig.~\ref{fig:intro_overview}(a), future-oriented cues can be particularly
valuable during search, transit, and approach, where anticipating subsequent
states facilitates goal-directed motion. In contrast, contact-rich stages such
as grasping and placement require stronger grounding in immediate interaction
feedback, making uncertain future predictions potentially less useful for
precise control. Moreover, even within the same execution stage, individual
future latents may differ in their relevance to the current control objective.
To empirically validate this intuition, we conduct a stage-wise analysis of
imagination utilization in Sec.~\ref{sec:rethinking_imagination_usage}. The
analysis reveals two consistent forms of non-uniformity in future utility:
\textbf{(i)} at the inter-progress level, the utility of imagined futures
varies substantially across execution stages as control demands change; and
\textbf{(ii)} at the intra-progress level, individual future latents
exhibit heterogeneous relevance within the same progress state. 

Building on these findings, we propose \textbf{ProWAM}, a
\textbf{Progress-Conditioned World Action Model} that introduces execution
progress as an explicit intermediate representation for adaptive imagination
utilization. As illustrated in Fig.~\ref{fig:intro_overview}(b), Vanilla VLAs
directly map observed context to actions without explicitly modeling future
dynamics, whereas standard WAMs augment action generation with imagined future
states but typically lack progress-aware regulation of how such information is
utilized. ProWAM instead explicitly models the evolving execution progress and
uses it to adaptively regulate imagined future information throughout action
generation. The resulting framework comprises two tightly coupled components:

\textbf{(1) Self-Supervised Dual-Temporal Progress Encoder (SS-DTPE).}
To obtain a progress representation that reflects both immediate interaction
status and accumulated task advancement, we propose SS-DTPE with a dual-temporal
design. Its Short-Term Action-Observation Encoder explicitly models language-conditioned
action--observation interactions, allowing the encoder to capture progress cues
revealed by the effect of the preceding action on the current observation.
Its Long-Term Progress Aggregator comprises a recurrent
Mamba~\cite{gu2023mamba} path followed by Progress Attention. The Mamba path
maintains compact historical progress memory to disambiguate visually similar
states at different execution stages, while Progress Attention enables the
memory-filtered progress representation query short-term action--observation
evidence and selectively absorb the latest task-relevant feedback. 
Beyond this architectural design, we introduce two self-supervised training
objectives that guide SS-DTPE to acquire progress awareness from the intrinsic
semantic progression and temporal ordering of demonstrations, without relying on
manual progress annotations. Together, these designs yield 
temporally coherent progress representations for subsequent modulation.

\textbf{(2) Hierarchical Progress-Conditioned Imagination Modulation (HPIM).}
Conditioned on the progress representation from SS-DTPE, we propose HPIM to
adapt imagination utilization to the evolving execution progress. Rather than
treating imagined futures with a fixed utilization pattern, HPIM regulates the
action-to-future attention pathway inside the MoT backbone, controlling how
future latents interact with the action stream through two complementary levels.  At the inter-progress level, the
Inter-Progress Global Gate adapts future utilization across execution stages
with different control demands. At the intra-progress level, the Intra-Progress
Future-Latent Relevance mechanism differentiates individual future latents
within the same progress state. These
two levels directly correspond to our empirical findings, capturing both
progress-dependent variations in future utility and latent-specific relevance.
Together, they turn fixed imagination usage into progress-conditioned
future utilization, enabling the action stream to adaptively exploit imagined
futures according to the progress state.

We conduct extensive evaluations on five simulation benchmarks, including LIBERO~\cite{liu2023libero}, RoboTwin2.0~\cite{chen2025robotwin}, VLABench~\cite{zhang2025vlabench}, RoboEval~\cite{wang2025roboeval}, and Mikasa-Robo~\cite{cherepanov2025memory}, covering general manipulation, robustness and generalization, progress-aware execution, and temporal memory. We further evaluate ProWAM on two real robotic platforms, Galaxea R1 Lite~\cite{galaxea} and AgileX Cobot
Magic~\cite{aloha}, under progress-aware long-horizon manipulation and generalization settings. As shown in Fig.~\ref{fig:intro_overview}(c), ProWAM consistently outperforms
strong VLA and WAM baselines,
demonstrating the effectiveness and generality of ProWAM.
In summary, our main contributions are as follows:
\begin{itemize}[leftmargin=0pt,itemsep=0pt,parsep=0pt,topsep=0pt]

    \item We identify two underexplored forms of non-uniformity in future
    utility: inter-progress variation across execution stages and
    intra-progress heterogeneity among future latents within the same progress
    state.

    \item We propose \textbf{SS-DTPE}, a self-supervised dual-temporal encoder
    that couples short-term action--observation evidence with long-term recurrent
    history to learn structured progress representations without manual
    annotations.

  \item We propose \textbf{HPIM}, a hierarchical progress-conditioned attention
modulation mechanism that  aligns imagination utilization with
execution progress through inter-progress global gating and intra-progress
future-latent relevance.

    \item Extensive experiments on five simulation benchmarks and two
    real-world robotic platforms demonstrate consistent gains over strong VLA
    and WAM baselines, particularly on progress-aware and long-horizon
    manipulation.
\end{itemize}

\vspace{-10pt}
\section{Related work}
\label{sec:rw}
\subsection{Vision-Language-Action Models}
Vision-Language-Action (VLA) models have become a prominent paradigm for
general-purpose robotic manipulation by grounding visual observations and
language instructions into executable actions~\cite{
li2025lion, lyu2025puma, zhu2025emosym, zhang2025falcon, li2026cogvla, zhu2026uniemo, zhou2026hiconagent, li2026consisvla, lyu2026personalalign, shao2026hats, li2025star, zhu2026delta}. Early large-scale VLAs adapt
pretrained vision-language representations~\cite{shao2024detecting, shao2023detecting, shao2019multi, xiao2026survey, ye2022unsupervised, wang2026affectagent, yuan2026coemogen, shen2024mome, li2024optimus} to visuomotor control, while
subsequent studies improve action generation, data efficiency, generalization,
and long-horizon reasoning
\cite{kim2025fine,black2024pi_0,intelligence2025pi_,song2025pd,
shukor2025smolvla,li2026cogvla,li2026semanticvla,shi2025memoryvla,peng2026survey, shao2025large}.

More recently, predictive VLAs have incorporated future-oriented information
into policy learning. Some methods explicitly generate future images or visual
foresight as intermediate guidance
\cite{zhao2025cot,lv2025f1,shen2026videovla}, while others model world knowledge
or latent future representations to capture anticipated state transitions
\cite{zhang2026dreamvla,xu2026futurevla}. These advances demonstrate the value of
predictive information for action generation and motivate World Action Models,
which more tightly couple future visual dynamics with robot actions. 

\vspace{-5pt}
\subsection{World Action Models}
\label{sec:related_wam}

World Action Models (WAMs) extend conventional robot policies by coupling
future visual modeling with action generation, allowing anticipated world
dynamics to provide predictive supervision for control. Early approaches
explore unified world--action modeling through autoregressive or joint
video--action generation
\cite{cen2025worldvla,bi2026motus,li2026causal}. Subsequent studies improve
future representations and physical grounding
\cite{li2026egowam,li2026wam4d}, or pursue more efficient model adaptation and
deployment
\cite{yuan2026fast,li2026light,jiang2026ckt}. Recent works further investigate
adaptive information gating and execution
\cite{huang2026noisegate,wang2026trust}, as well as high-level planning,
long-term memory, and task-progress modeling for complex manipulation
\cite{zhu2026dswam,wang2026dim}.

These advances substantially broaden how imagined future information is modeled
and incorporated into WAM-based control. Recent studies have further explored
adaptive mechanisms, including per-latent information gating
\cite{huang2026noisegate}, future--reality verification for adaptive execution
\cite{wang2026trust}, and memory-augmented progress modeling for long-horizon
manipulation \cite{wang2026dim}. Nevertheless, these approaches mainly adapt
denoising schedules, execution horizons, or temporal memory,
rather than explicitly translating a structured execution-progress
representation into the regulation of imagined future latents. In contrast,
ProWAM learns execution progress explicitly and uses it to hierarchically
modulate future-latent attention at both inter-progress and intra-progress
levels, directly coupling the evolving task progress with imagination
utilization.

\vspace{-5pt}
\subsection{Progress Modeling for Robotic Manipulation}
\label{sec:related_progress}

Recent studies have explored explicit progress modeling for long-horizon
robotic manipulation. See, Plan, Rewind~\cite{dai2026see} represents task
progress through spatial milestones for execution monitoring and recovery,
while ProgVLA~\cite{kim2026progvla} learns progress estimates to improve policy
learning. RARM~\cite{yang2026rarm} leverages task advancement for progress-aware
reward modeling, and HOST~\cite{chen2026robots} aligns human and robot
trajectories in a shared task-progress space for skill acquisition. In the WAM
setting, TempoWAM~\cite{ye2026rethink} uses a recurrent progress monitor to
adapt execution and replanning. These studies demonstrate the value of progress
awareness for temporally coherent manipulation.

ProWAM differs in both progress representation and utilization. SS-DTPE learns
structured execution progress from short-term action--observation evidence and
long-term recurrent history without manual progress annotations. Rather than
using progress primarily for monitoring, policy learning, reward modeling, or
execution scheduling, ProWAM further conditions HPIM on the learned progress
state to hierarchically modulate imagined future latents. This directly couples
execution-progress modeling with adaptive imagination utilization.
    
\begin{figure*}[t]
    \centering
    \includegraphics[width=1.0\textwidth]{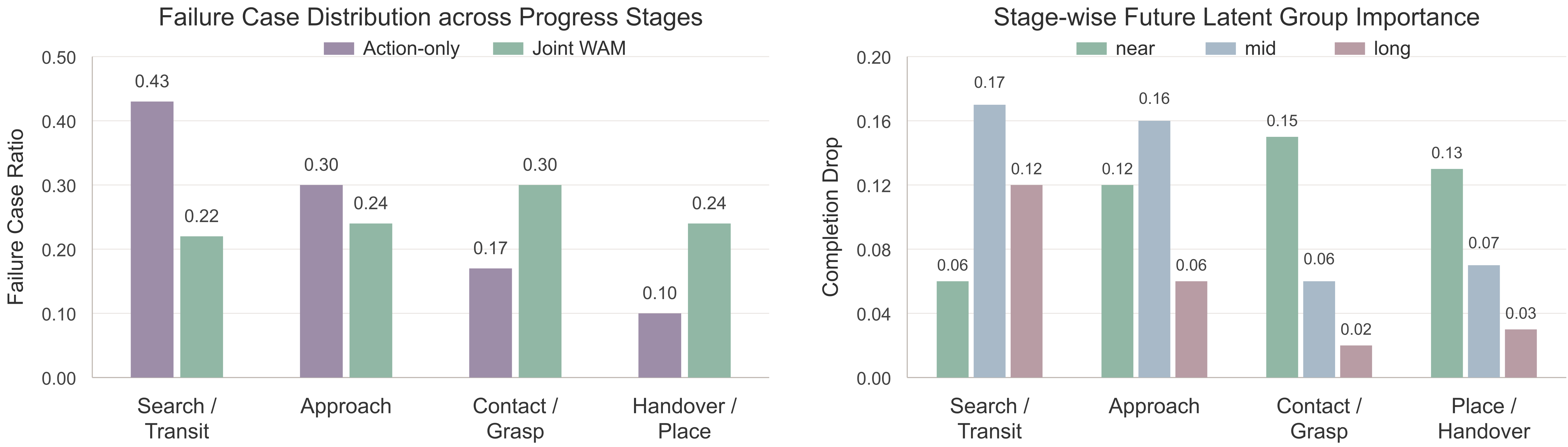}
    \vspace{-15pt}
\caption{
\textbf{Empirical Motivation for Progress-Conditioned Imagination Usage.}
\textbf{Left}: Failure-case distribution across execution progress. The action-only variant tends to produce more failures in pre-contact stages such as search/transit and approach, suggesting that future visual latents can provide useful foresight before contact-rich interaction.
\textbf{Right:} Sensitivity to perturbed future latent groups. Different progress stages show different sensitivity to near-, mid-, and long-horizon future latents, indicating that future information is not uniformly useful across the imagined sequence.
}
\vspace{-10pt}
    \label{fig:imagination_heterogeneity}
\end{figure*}

\vspace{-5pt}
\section{Problem Formulation}
\label{sec:problem_formulation}
\subsection{Preliminaries: Video--Action Generation in WAMs}
\label{sec:prelim_wam}
At timestep \(t\), the robot observes the current visual input \(O_t\) and task
instruction \(L\), and predicts an action chunk:
\begin{equation}
    A_t = \{a_t,a_{t+1},\ldots,a_{t+H_a-1}\},
    \label{eq:action_chunk}
\end{equation}
where \(a_t\) is a low-level robot action and \(H_a\) is the action horizon. For
single-arm manipulation, \(a_t\) is typically a 7-DoF command consisting of
end-effector translation, rotation, and gripper state.
A standard visuomotor policy \(\pi_{\theta}\)  directly predicts this action chunk from the
current observation-instruction context:
\begin{equation}
    A_t \sim \pi_{\theta}(\cdot \mid O_t,L).
    \label{eq:visuomotor_policy}
\end{equation}

Although this direct formulation is efficient, it does not explicitly model how
the scene may evolve after the current decision. This limitation is important in
manipulation, where action generation often benefits from anticipating object
motion, robot--object contact, and task-relevant state transitions. World Action
Models (WAMs) address this issue by coupling action generation with future visual
prediction. We focus on the joint WAM paradigm, where future visual
latents and action tokens are modeled within a shared generative process.
Specifically, let \(Z_t^v=\{z_{t+1}^v,\ldots,z_{t+H_v}^v\}\) denote the future
visual latent sequence over a video horizon \(H_v\), where \(z_{t+k}^v\) is the
latent visual state at the \(k\)-th future step. Given \(O_t\) and \(L\), a joint
WAM \(f_{\theta}\) predicts both future visual latents \(\widehat{Z}_t^v\) and actions \(\widehat{A}_t\):
\begin{equation}
    (\widehat{Z}_t^v,\widehat{A}_t)
    =
    f_{\theta}(O_t,L).
    \label{eq:joint_wam_prediction}
\end{equation}

A common implementation adopts a diffusion or flow-based
Mixture-of-Transformers (MoT)~\cite{liang2024mixture} backbone. Unlike a
homogeneous transformer, MoT preserves modality-specific computation for video
and action tokens while allowing them to communicate through shared attention.
Let \(H_v^l\) and \(H_a^l\) denote the video-token and action-token hidden states
at layer \(l\). A MoT block is described as:
\begin{equation}
    (H_v^{l+1},H_a^{l+1})
    =
    \mathcal{T}_{\theta}^{l}
    \left(
    \Phi_v^l(H_v^l),
    \Phi_a^l(H_a^l);
    O_t,L
    \right),
    \label{eq:mot_update}
\end{equation}
where \(\Phi_v^l\) and \(\Phi_a^l\) are modality-specific video and action
experts, and \(\mathcal{T}_{\theta}^{l}\) is the shared video--action attention
block. This coupling enables action tokens to access spatio-temporal cues from
imagined future visual tokens while preserving an action-specialized computation
path.
During training, the joint WAM is optimized with rectified flow-matching losses
for future visual prediction and action generation:
\begin{equation}
    \mathcal{L}_{\mathrm{WAM}}
    =
    \mathcal{L}_{\mathrm{act}}
    +
    \mathcal{L}_{\mathrm{vid}},
    \label{eq:wam_loss}
\end{equation}
where \(\mathcal{L}_{\mathrm{act}}\) supervises action-token denoising and
\(\mathcal{L}_{\mathrm{vid}}\) supervises future-video-latent denoising.

\subsection{Rethinking Imagination Usage}
\label{sec:rethinking_imagination_usage}

\noindent\textbf{Does Task Progress Affect Imagination Usage?}
WAMs make future visual latents accessible to the action stream through shared video-action modeling. While this formulation provides the policy with imagined future dynamics, it also raises a natural question: once future latents are available, should they influence action generation in the same way throughout the entire execution? Long-horizon manipulation is temporally structured. A robot may first search for a target, approach the object, establish contact, and then perform fine-grained grasping, handover, or placement. These progress regimes impose different control requirements. Pre-contact stages often benefit from foresight about target direction, object-reaching trends, and upcoming state transitions, whereas contact-rich stages depend more on current visual evidence, proprioceptive feedback, and precise local control. This suggests that the usefulness of imagined futures may vary with execution progress.

\vspace{3pt}
\noindent\textbf{Experimental Setup.}
To obtain empirical evidence for this intuition, we conduct five real-world long-horizon manipulation tasks on the Galaxea R1 Lite platform: \textbf{Drawer Manipulation, Plate Handover, Markers Collection, Plate Lemon Apple, and Object Packing}. Each task is evaluated with 100 trials, resulting in 500 trials in total. These tasks cover diverse execution progress, including target search, object approach, contact establishment, collection, handover, and fine-grained placement.
We instantiate a WAM following the joint variant of Fast-WAM~\cite{yuan2026fast}. The model is built upon the video Diffusion Transformer from Wan2.2-5B~\cite{wan2025wan}, reusing its pretrained text encoder and video VAE. We introduce an action expert DiT for action chunk generation and organize the full model as a Mixture-of-Transformers (MoT) architecture with shared attention.

We consider two diagnostic settings. (1)~To probe whether future latents have different overall utility across progress regimes, we compare the joint WAM with an action-only variant, where the action stream is prevented from accessing future visual latents while the action backbone, language conditioning, observation encoder, and action objective are kept unchanged. We then categorize each failed trial by the progress stage where the failure occurs. (2)~To probe whether different future latents contribute equally within a progress regime, we perturb future latents by horizon groups, including near-, mid-, and long-horizon groups, and measure the resulting stage completion drop.

\vspace{3pt}
\noindent\textbf{Finding 1: Future Latents Show Progress-Dependent Overall Utility.}
As shown in Fig.~\ref{fig:imagination_heterogeneity} (Left), the action-only variant tends to produce a larger fraction of failures in search/transit and approach stages. In comparison, the joint WAM reduces the relative frequency of these early-stage failures, while the remaining failures are more often associated with contact-rich stages. This pattern suggests that future visual latents are particularly helpful before physical interaction, where action generation can benefit from foresight for target-directed motion and object approach. Once the robot enters contact, grasping, or placement, fine-grained local control becomes more critical. 

\vspace{3pt}
\noindent\textbf{Finding 2: Future Latents Show Progress-Dependent Local Relevance.}
The perturbation analysis in Fig.~\ref{fig:imagination_heterogeneity} (Right) further suggests  that different future horizons contribute differently across progress regimes. Pre-contact stages, such as search/transit and approach, are more sensitive to mid- and long-horizon future latents, indicating that imagined motion trends and target-reaching directions can be useful. In contrast, contact-rich stages such as grasping, placement, and handover are more sensitive to near-future latents, while distant imagined states may become less reliable or less relevant for immediate control. This observation suggests that, even under a similar execution progress, future latents should not be treated as uniformly useful; their relevance depends on the control information.

\vspace{3pt}
\noindent\textbf{Key Discovery.}
The diagnostic results highlight \textbf{execution progress as the key intermediate signal} for understanding imagination usage. Since the usefulness of future latents varies along the execution trajectory, an online progress representation is needed to determine where the robot is within the current task. Without such progress information, future latents can only be exposed to the action stream as a fixed spatio-temporal context.

Taken together, the diagnostic analysis reveals two complementary forms of
non-uniformity in future utility. \textbf{(1) Inter-progress variation.}
The utility of imagined futures varies across execution stages as control
demands evolve. \textbf{(2) Intra-progress variation.} Within the same progress
state, individual future latents exhibit heterogeneous relevance.
\vspace{-10pt}
\section{Method}
\label{sec:method}
Based on the above analysis, we propose \textbf{ProWAM}, a
\textbf{Pro}gress-conditioned \textbf{W}orld \textbf{A}ction \textbf{M}odel
that adaptively utilizes imagined future latents for action generation. ProWAM
consists of two core modules: a \textbf{S}elf-\textbf{S}upervised
\textbf{D}ual-\textbf{T}emporal \textbf{P}rogress \textbf{E}ncoder
(\textbf{SS-DTPE}) for online progress estimation, and
\textbf{H}ierarchical \textbf{P}rogress-Conditioned
\textbf{I}magination \textbf{M}odulation (\textbf{HPIM}) for progress-aware
future-latent utilization.
We first describe the overall framework of ProWAM in
Sec.~\ref{sec:overall_framework}. We then introduce SS-DTPE and HPIM in
Sec.~\ref{sec:progress_encoding} and Sec.~\ref{sec:imagination_modulation},
respectively. Finally, we detail the training and inference procedure in
Sec.~\ref{sec:training_inference}.

\subsection{Overall Framework}
\label{sec:overall_framework}
\begin{figure*}[t]
    \centering
    \includegraphics[width=\textwidth]{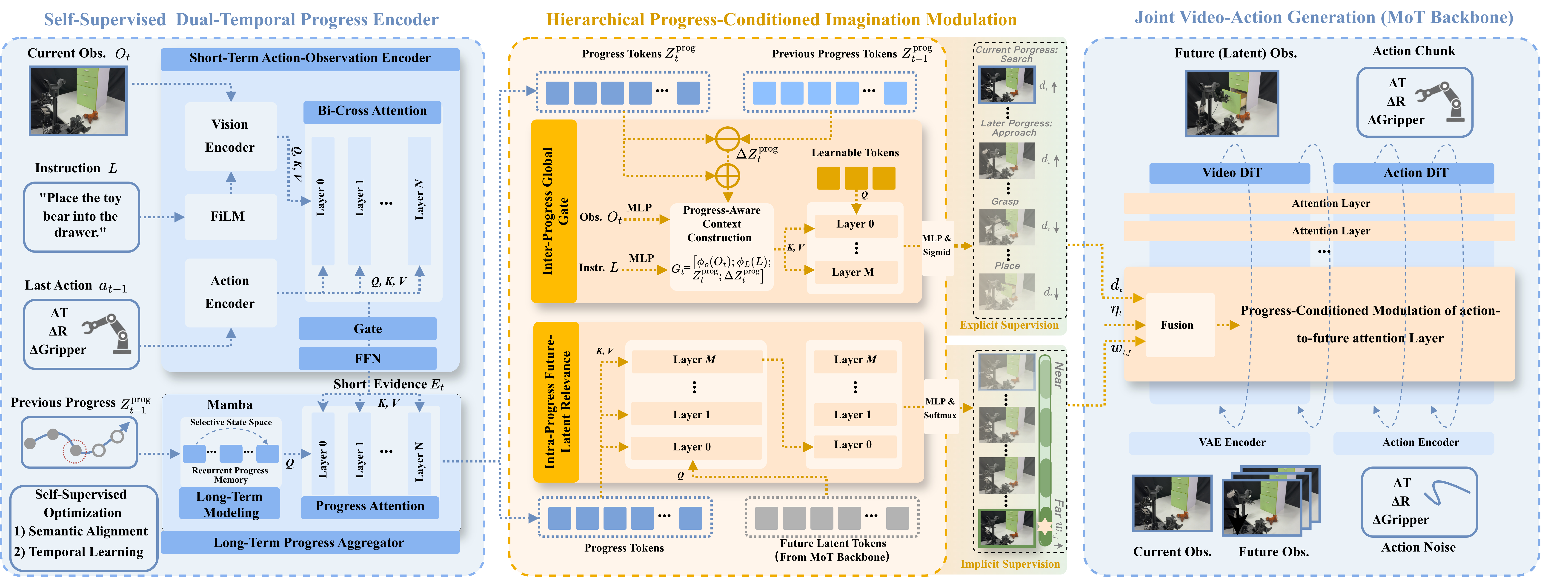}
    \vspace{-10pt}
    \caption{\textbf{Overview of ProWAM.}
ProWAM comprises a Self-Supervised Dual-Temporal Progress Encoder (SS-DTPE)
and Hierarchical Progress-Conditioned Imagination Modulation (HPIM).
SS-DTPE integrates short-term action--observation evidence with long-term
execution history to encode the current progress state. Conditioned on this
representation, HPIM jointly models an inter-progress global gate \(d_t\) and
intra-progress future-latent relevance \(\omega_{t,f}^{l}\), and combines them
with the layer-wise coefficient \(\eta_l\) to obtain the modulation weight
\(\beta_{t,f}^{l}\). The resulting modulation is applied to the action-to-future attention pathway,
enabling adaptive utilization of imagined future information.}
    \label{fig:framework}
    \vspace{-8pt}
\end{figure*}
As shown in Fig.~\ref{fig:framework}, \textbf{ProWAM} follows a
progress-first design for adaptive future utilization in World Action Models.
Given the current observation \(O_t\), task instruction \(L\), and the previously
executed action \(a_{t-1}\), the Self-Supervised Dual-Temporal Progress Encoder
(\textbf{SS-DTPE}) represents execution progress with a set of latent progress
tokens \(Z_t^{\mathrm{prog}}\). These tokens are learned progress state slots
rather than predefined phase labels. They serve as long-term progress memory,
which is maintained through a Mamba-based progress path, while the current pair
\((a_{t-1},O_t)\) provides short-term action-observation feedback about the
latest execution outcome. At each timestep, SS-DTPE updates the progress state by
selectively absorbing this short-term feedback into the accumulated progress
memory:
\begin{equation}
Z_t^{\mathrm{prog}}
=
\operatorname{SS\mbox{-}DTPE}
\left(
O_t,
a_{t-1},
L,
Z_{t-1}^{\mathrm{prog}}
\right),
\label{eq:overall_progress_update}
\end{equation}
where \(Z_t^{\mathrm{prog}}\) is the progress representation at
timestep \(t\).

To endow SS-DTPE with progress estimation capability, we avoid relying on
fine-grained manual annotations, which are costly to obtain and often
ambiguous for long-horizon manipulation trajectories. Instead, we exploit the
intrinsic structure of demonstration data and formulate progress extraction as
self-supervised progress representation learning. Specifically, SS-DTPE uses the
temporal ordering of demonstrations and the instruction-level task structure as
natural supervision signals, encouraging progress tokens to evolve consistently
with task execution while remaining aligned with semantic task advancement. Once
trained, SS-DTPE can infer the current progress representation online from
\(O_t\), \(a_{t-1}\), \(L\), and \(Z_{t-1}^{\mathrm{prog}}\). We detail its
dual-temporal architecture and self-supervised objectives in
Sec.~\ref{sec:progress_encoding}.

We build ProWAM upon the formulation introduced in
Sec.~\ref{sec:prelim_wam}. Let \(H_v^l\) and \(H_a^l\) denote the future-latent
and action-token hidden states at the \(l\)-th MoT layer, respectively. Given the
progress tokens \(Z_t^{\mathrm{prog}}\), the Hierarchical Progress-Conditioned
Imagination Modulation module (\textbf{HPIM}) predicts layer-wise importance
signals for utilization:
\begin{equation}
\Omega_t^l
=
\operatorname{HPIM}^{l}
\left(
Z_t^{\mathrm{prog}},
H_v^l,
H_a^l
\right).
\label{eq:overall_hpim_signal}
\end{equation}
Motivated by the observation that imagination utility varies both across
execution progress and across predicted future latents, HPIM models future
utilization from two complementary perspectives:
\begin{equation}
\Omega_t^l
=
\alpha_t^l \cdot \omega_t^l,
\qquad
\alpha_t^l \in \mathbb{R}_{+},
\quad
\omega_t^l \in \mathbb{R}_{+}^{F}.
\label{eq:overall_hierarchical_modulation}
\end{equation}
Here, \(F\) is the number of future latent tokens at layer \(l\). The scalar
\(\alpha_t^l\) captures inter-progress regulation, estimating the overall
reliance on imagined futures under the current progress state. The vector
\(\omega_t^l\) captures intra-progress regulation, assigning different
importance to individual future latents within the predicted sequence. Their
product \(\Omega_t^l\) provides a progress-conditioned importance signal.

Rather than using progress as an auxiliary feature after action prediction, HPIM
directly modulates the MoT attention path from action-token queries to
future-latent keys. Let \(C_{a\rightarrow v}^{l}(i,f)\) denote the original
attention logit from the \(i\)-th action token to the \(f\)-th future latent
token. We inject the progress-conditioned importance into this attention path:
\begin{equation}
\widetilde{C}_{a\rightarrow v}^{l}(i,f)
=
C_{a\rightarrow v}^{l}(i,f)
+
\log\left(\Omega_{t,f}^{l}+\epsilon\right),
\label{eq:overall_attention_modulation}
\end{equation}
where \(\epsilon\) is a small constant for numerical stability. The
progress-modulated MoT update is abstractly written as:
\begin{equation}
\left(
H_v^{l+1},
H_a^{l+1}
\right)
=
\operatorname{MoT}^{l}
\left(
\Phi_v^l(H_v^l),
\Phi_a^l(H_a^l);
O_t,
L,
\widetilde{C}_{a\rightarrow v}^{l}
\right),
\label{eq:overall_modulated_mot}
\end{equation}
where \(\Phi_v^l\) and \(\Phi_a^l\) are the modality-specific video and action
experts. In this way, progress information directly adjusts how strongly
imagined future latents are attended by the action stream, enabling ProWAM to
selectively strengthen or suppress future information instead of uniformly
trusting or discarding imagination.

After the progress-modulated WAM backbone, the final action representation is
used to predict the action chunk:
\begin{equation}
\widehat{A}_t
=
\operatorname{ActHead}
\left(
H_a^{\mathrm{out}}
\right),
\qquad
\widehat{A}_t
=
\{\hat a_t,\ldots,\hat a_{t+H_a-1}\},
\label{eq:overall_action_output}
\end{equation}
where \(H_a^{\mathrm{out}}\) denotes the final action representation produced by
the modulated WAM backbone. The future-latent stream follows the standard WAM
training objective, while ProWAM focuses on progress-conditioned modulation of
its attention influence on the action stream.

\subsection{Self-Supervised Dual-Temporal Progress Encoder}
\label{sec:progress_encoding}
A prerequisite for progress-conditioned future utilization is an explicit
representation of where the robot is within the current task execution.
However, estimating execution progress in closed-loop robotic manipulation is
nontrivial. The current observation alone is often insufficient, since visually
similar states may imply different progress depending on the action that has just
been executed. Thus, progress estimation should capture short-term
action-observation feedback. Meanwhile, long-horizon tasks require memory of
past execution evidence, as similar local configurations may recur at different
stages of a trajectory.

Motivated by this dual-temporal requirement, we propose a \textbf{Dual-Temporal Progress Encoder}, which couples a \textbf{Short-Term Action-Observation Encoder} for real-time feedback extraction with a \textbf{Long-Term Progress Aggregator} for recurrent progress modeling that maintains an evolving progress memory by selectively absorbing task-relevant action-observation feedback over time.

\vspace{2pt}
\noindent\textbf{Short-Term Action-Observation Encoder.}
Given the current observation \(O_t\), the previously executed action
\(a_{t-1}\), and the task instruction \(L\), the short-term encoder extracts
local feedback evidence that reflects how the scene responds to the latest
action. We first encode \(O_t\) into observation tokens \(X_t\) and project
\(a_{t-1}\) into action tokens \(U_{t-1}\). To make the visual evidence
task-aware, we apply instruction-conditioned Feature-wise Linear Modulation
(FiLM):
\begin{equation}
\begin{gathered}
X_t = \phi_{\mathrm{obs}}(O_t), 
U_{t-1} = \phi_{\mathrm{act}}(a_{t-1}), \\
\widetilde{X}_t
=
X_t \odot (1+\gamma_L) + \beta_L .
\end{gathered}
\label{eq:short_term_tokens}
\end{equation}
Here, \(\phi_{\mathrm{obs}}\) and \(\phi_{\mathrm{act}}\) denote the observation
and action encoders, respectively. The FiLM parameters \(\gamma_L\) and
\(\beta_L\) are text-conditioned scale and shift vectors, and \(\odot\) denotes element-wise multiplication.

To explicitly capture the feedback relation between the executed action and the
current observation, we use bidirectional cross-attention:
\begin{equation}
\begin{aligned}
F_t^{o\leftarrow a}
&=
\widetilde{X}_t
+
\operatorname{Attn}
\left(
Q=\widetilde{X}_t,\,
K=U_{t-1},\,
V=U_{t-1}
\right), \\
F_t^{a\leftarrow o}
&=
U_{t-1}
+
\operatorname{Attn}
\left(
Q=U_{t-1},\,
K=\widetilde{X}_t,\,
V=\widetilde{X}_t
\right).
\end{aligned}
\label{eq:short_term_cross_attention}
\end{equation}
The first stream injects action-conditioned cues into the observation tokens,
highlighting visual evidence related to the previous action. The second stream
queries the current observation with action tokens, summarizing the visual
outcome associated with the executed action. These two directions provide
complementary views of short-term execution feedback.
We then fuse the two interaction streams into short-term action-observation
evidence:
\begin{equation}
E_t
=
\psi_e
\left(
\left[
\psi_o(F_t^{o\leftarrow a});
\psi_a(F_t^{a\leftarrow o})
\right]
\right),
\label{eq:short_term_evidence}
\end{equation}
where \(\psi_o\), \(\psi_a\), and \(\psi_e\) are lightweight projection modules.
The resulting evidence tokens \(E_t\) encode task-conditioned and action-aware
feedback at the current step, which will be selectively absorbed by the
long-term progress aggregator.

\vspace{2pt}
\noindent\textbf{Long-Term Progress Aggregator.}
While the short-term encoder provides local action-observation evidence \(E_t\),
execution progress also requires a persistent memory over the past trajectory.
We use a Mamba-based progress aggregator for this purpose, since its selective
state-space mechanism supports efficient long-sequence modeling and
input-dependent memory updates. This allows the progress tokens
\(Z_t^{\mathrm{prog}}\) to preserve slowly evolving task context while
selectively incorporating newly observed action-observation evidence when it is
relevant to progress estimation.
Following Mamba~\cite{gu2023mamba}, we model the progress memory with an
input-dependent selective state-space recurrence. Given the previous progress
tokens, we first obtain a normalized progress representation and generate the
selective transition parameters:
\begin{equation}
\begin{gathered}
R_{t-1}
=
\operatorname{LayerNorm}
\left(
Z_{t-1}^{\mathrm{prog}}
\right), \\
\Theta_t
=
\operatorname{Select}_{\mathrm{prog}}
\left(
R_{t-1}
\right)
=
\{\Delta_t,\bar{A}_t,\bar{B}_t,C_t,g_t\}.
\end{gathered}
\label{eq:progress_selective_params}
\end{equation}
Here, \(\operatorname{Select}_{\mathrm{prog}}\) is the input-conditioned
parameter generator of the progress Mamba block. The set \(\Theta_t\) controls
how the progress memory is updated: \(\Delta_t\) determines the adaptive
timescale, \(\bar{A}_t\) and \(\bar{B}_t\) are the discretized state transition
and input parameters, \(C_t\) is the readout projection, and \(g_t\) is the
output gate.
The recurrent progress memory is then updated as follows:
\begin{equation}
\begin{gathered}
h_t^{\mathrm{prog}}
=
\bar{A}_t h_{t-1}^{\mathrm{prog}}
+
\bar{B}_t R_{t-1}, \\
\widetilde{P}_t
=
Z_{t-1}^{\mathrm{prog}}
+
\operatorname{Linear}
\left(
C_t h_t^{\mathrm{prog}}
\odot
\sigma(g_t)
\right).
\end{gathered}
\label{eq:progress_memory_update}
\end{equation}
where \(h_t^{\mathrm{prog}}\) denotes the recurrent memory state,
\(\widetilde{P}_t\) is the memory-filtered progress representation, and
\(\sigma(\cdot)\) denotes the SiLU activation. This update preserves accumulated
trajectory context before incorporating current evidence.

To inject newly observed execution feedback, we introduce
\textbf{Progress Attention}. The memory-filtered progress representation
\(\widetilde{P}_t\) queries the short-term evidence tokens \(E_t\), allowing the
progress state to selectively absorb task-relevant action-observation feedback:
\begin{equation}
Z_t^{\mathrm{prog}}
=
\operatorname{FFN}
\left(
\widetilde{P}_t
+
\operatorname{Attn}
\left(
Q=\widetilde{P}_t,
K=E_t,
V=E_t
\right)
\right).
\label{eq:progress_evidence_attention}
\end{equation}
This design separates long-term memory maintenance from current evidence
absorption: the Mamba path preserves historical progress context, while
Progress-Evidence Attention selectively integrates the latest
action-observation feedback into the progress tokens.

\begin{figure*}[t]
    \centering
    \includegraphics[width=\textwidth]{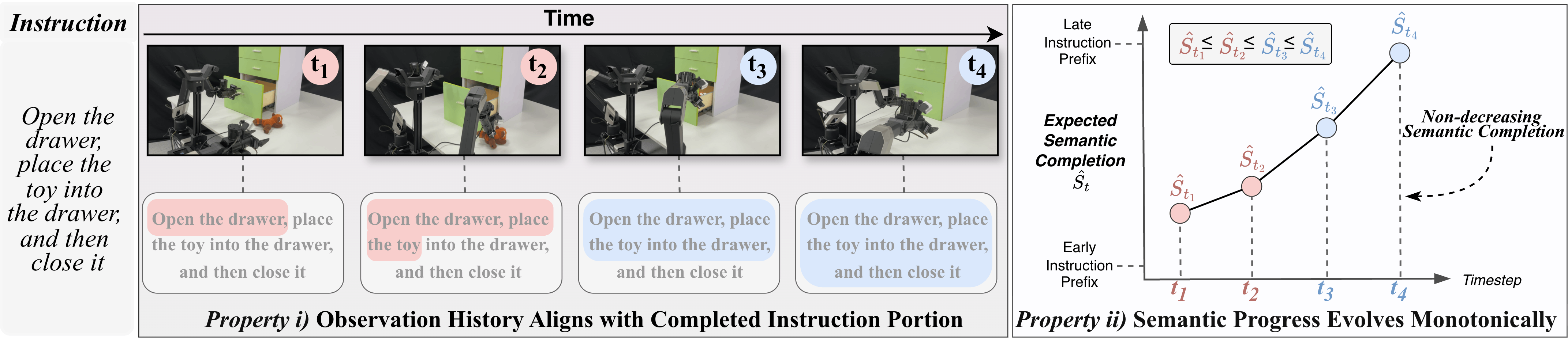}
    \vspace{-10pt}
\caption{
\textbf{Self-Supervised Semantic Progress Alignment.}
\textbf{Left (Property i):} As execution progresses, the observation history
aligns with increasingly complete instruction prefixes, which serve as latent
semantic progress states.
\textbf{Right (Property ii):} The expected semantic completion is constrained
to evolve monotonically along successful trajectories.
}
    \label{fig:semantic_progress_alignment}
    \vspace{-8pt}
\end{figure*}

\vspace{2pt}
\noindent\textbf{Self-Supervised Training Objectives.}
To endow SS-DTPE with progress-aware representation capability without costly
and ambiguous manual progress annotations, we exploit the intrinsic structure of
demonstration trajectories. Specifically, successful demonstrations provide two
natural sources of self-supervision: instruction-level task structure and
temporal execution order. Based on these cues, we train the progress tokens with
two complementary objectives:

\noindent\textbf{1) Semantic Progress Alignment.}
Instead of relying on unavailable step-level progress annotations, we derive self-supervision from two structural properties in Fig.~\ref{fig:semantic_progress_alignment}: \textbf{i)} The observation history up to timestep \(t\) should correspond to the prefix of the instruction that has been completed so far. \textbf{ii)} Semantic progress should evolve monotonically with the visual trajectory. 
Thus, we treat instruction prefixes as latent progress states and train progress tokens to infer the completed instruction portion.

\textbf{For the property i)}, given the instruction tokens \(C=\{c_1,\ldots,c_M\}\), we construct a sequence of prefix representations \(\{p_j\}_{j=1}^{M}\), where each \(p_j=\frac{1}{j}\sum_{i=1}^{j}W_c c_i\) summarizes the instruction semantics up to the \(j\)-th token. These prefix embeddings define a semantic progress axis from early task intent to later task completion.
At timestep \(t\), the Dual-Temporal Progress Encoder outputs progress tokens \(Z_t^{\mathrm{prog}}\). We match them against all prefix representations to predict a semantic progress distribution:
\begin{equation}
\pi_t
=
\operatorname{Softmax}_{j=1}^{M}
\left(
\frac{
\cos\left(g(Z_t^{\mathrm{prog}}),p_j\right)
}{
\tau_p
}
\right),
\end{equation}
where \(g(\cdot)\) is a lightweight projection head, \(\tau_p\) is a temperature parameter, and \(\pi_t\) denotes the predicted distribution over completed instruction prefixes.
Following prior work~\cite{zhao2026tapsampling}, we use normalized temporal progress as a weak linear prior for instruction completion. For a trajectory of length \(T\) with \(t\in\{0,\ldots,T-1\}\), we define \(r_t=t/(T-1)\) and map it to an instruction-prefix center \(\mu_t=1+r_t(M-1)\). Since temporal progress only provides a coarse estimate of semantic completion, we construct a Gaussian soft target over prefix positions as follows:
\begin{equation}
y_{t,j}
=
\frac{
\exp\left(-{(j-\mu_t)^2}/{2\sigma^2}\right)
}{
\sum_{k=1}^{M}
\exp\left(-{(k-\mu_t)^2}/{2\sigma^2}\right)
},
\quad j=1,\ldots,M,
\end{equation}
where \(\sigma\) controls the width of the Gaussian soft target. We optimize prefix alignment with a soft cross-entropy loss across the valid timestep \(\mathcal V\):
\begin{equation} \mathcal{L}_{\mathrm{sem}} = -\frac{1}{|\mathcal V|} \sum_{t\in\mathcal V} \sum_{j=1}^{M} y_{t,j}\log \pi_{t,j}, \end{equation}

\textbf{For the property ii)}, we further impose an order consistency constraint on the predicted semantic progress. Since execution should move forward along a successful trajectory, the semantic completion at an earlier timestep should not exceed that at a later timestep. Given the predicted prefix distribution \(\pi_t\), we use its expectation \(\hat{s}_t\) as the semantic completion position. We then penalize violations of this temporal ordering over adjacent timestep pairs \(\mathcal P\):
{\setlength{\abovedisplayskip}{2pt}
 \setlength{\belowdisplayskip}{2pt}
\begin{equation}
\begin{gathered}
\hat{s}_t
=
\mathbb{E}_{j\sim\pi_t}[j]
=
\sum_{j=1}^{M} j\pi_{t,j}, \\
\mathcal{L}_{\mathrm{ord}}
=
\frac{1}{|\mathcal P|}
\sum_{(t,t+1)\in\mathcal P}
\max
\left(
0,
\hat{s}_t-\hat{s}_{t+1}
\right).
\end{gathered}
\label{eq:order_loss}
\end{equation}}%
\noindent\textbf{2) Temporal Structure Learning.}
In addition to semantic progress alignment, we introduce a local temporal
contrastive objective to structure the progress space. This objective encourages
neighboring execution states to remain close while separating temporally distant
or semantically different states. For each valid timestep \(t\), we define:
\begin{equation}
\begin{gathered}
\mathbf{u}_t
=
\operatorname{Norm}
\left(
g_{\mathrm{temp}}(Z_t^{\mathrm{prog}})
\right), \\
\mathcal{L}_{\mathrm{temp}}
=
-\frac{1}{|\mathcal V|}
\sum_{t\in\mathcal V}
\log
\frac{
\exp(\mathbf{u}_t^\top \mathbf{u}_{t^+}/\tau_c)
}{
\sum_{k\in\mathcal B_t}
\exp(\mathbf{u}_t^\top \mathbf{u}_{k}/\tau_c)
}.
\end{gathered}
\label{eq:temporal_loss}
\end{equation}
Here, \(g_{\mathrm{temp}}(\cdot)\) is a temporal projection head, \(t^+\) is a
neighboring timestep from the same trajectory, \(\mathcal B_t\) denotes the
valid candidate set for timestep \(t\), and \(\tau_c\) is the temperature.
The overall self-supervised objective of SS-DTPE is:
\begin{equation}
\mathcal{L}_{\mathrm{SS\mbox{-}DTPE}}
=
\mathcal{L}_{\mathrm{sem}}
+
\lambda_{\mathrm{ord}}\mathcal{L}_{\mathrm{ord}}
+
\lambda_{\mathrm{temp}}\mathcal{L}_{\mathrm{temp}}.
\label{eq:ss_dtpe_loss}
\end{equation}
where 
\(\lambda_{\mathrm{ord}}\) and \(\lambda_{\mathrm{temp}}\) are balancing
hyperparameters.

\subsection{Hierarchical Progress-Conditioned Imagination Modulation}
\label{sec:imagination_modulation}

Given the progress representation learned by SS-DTPE, we next describe how it is
used to regulate imagination inside a joint video--action WAM. The analysis in
Sec.~\ref{sec:rethinking_imagination_usage} shows that the utility of imagined
future latents varies at two levels. Motivated by these observations, we introduce
\textbf{Hierarchical Progress-Conditioned Imagination Modulation}
(\textbf{HPIM}), which modulates future-latent utilization from both
inter-progress and intra-progress perspectives. 

We apply HPIM to the MoT attention path from action-token queries to
future-latent keys. Let
\(H_v^l=\{h_{v,f}^l\}_{f=1}^{F}\) and \(H_a^l\) denote the future-latent and
action-token hidden states at the \(l\)-th MoT layer, respectively, where \(F\)
is the number of future latent tokens. HPIM uses the progress tokens
\(Z_t^{\mathrm{prog}}\) produced by SS-DTPE as the progress condition and
generates a token-wise coefficient
\(\beta_t^l=\{\beta_{t,f}^l\}_{f=1}^{F}\) to control how imagined future latents
influence the action stream.

\vspace{2pt}
\noindent\textbf{Inter-Progress Global Gate.}
The inter-progress component estimates the stage-level reliance on imagined
futures under the current execution progress. To capture this
progress-dependent reliance, we construct a progress-aware context from the
SS-DTPE progress state:
\begin{equation}
\begin{gathered}
\Delta Z_t^{\mathrm{prog}}
=
Z_t^{\mathrm{prog}}
-
Z_{t-1}^{\mathrm{prog}}, \\
G_t
=
[
\phi_o(O_t);
\phi_L(L);
Z_t^{\mathrm{prog}};
\Delta Z_t^{\mathrm{prog}}
].
\end{gathered}
\label{eq:hpim_global_context}
\end{equation}
Here, \(\Delta Z_t^{\mathrm{prog}}\) describes the recent change of the progress
state, complementing the absolute progress representation
\(Z_t^{\mathrm{prog}}\). The functions \(\phi_o(\cdot)\) and \(\phi_L(\cdot)\)
are lightweight projections for the observation and instruction, respectively,
and \([\cdot]\) denotes token concatenation.

A learnable gate query aggregates this progress-aware context and predicts a
scalar global gate:
\begin{equation}
\begin{gathered}
h_t^{g}
=
\operatorname{Attn}
\left(
Q=q^{g},
K=G_t,
V=G_t
\right), \\
d_t
=
\sigma
\left(
\operatorname{MLP}_{g}
\left(
h_t^{g}
\right)
\right),
\qquad
d_t\in[0,1].
\end{gathered}
\label{eq:hpim_global_gate}
\end{equation}
where \(q^{g}\) is the learnable query, \(h_t^{g}\) is the aggregated
feature, and \(d_t\) represents the global reliance on imagined
futures. A larger \(d_t\) activates more selective use of future latents, while
a smaller \(d_t\) makes the modulation more conservative.

\vspace{2pt}
\noindent\textbf{Intra-Progress Future-Latent Relevance.}
The global gate determines the overall reliance on imagination, but it does not
distinguish which future latents are useful within the predicted sequence. To
capture intra-progress heterogeneity, HPIM further estimates token-wise future
relevance conditioned on the SS-DTPE progress tokens.
At the \(l\)-th MoT layer, future-latent tokens first query the progress state:
\begin{equation}
\begin{gathered}
U_t^l
=
\operatorname{Attn}
\left(
Q=H_v^l,
K=Z_t^{\mathrm{prog}},
V=Z_t^{\mathrm{prog}}
\right), \\
\bar{H}_v^l
=
H_v^l + U_t^l .
\end{gathered}
\label{eq:hpim_progress_conditioned_future}
\end{equation}
Here, \(U_t^l\) injects progress-aware context into the future-latent stream, and
\(\bar{H}_v^l\) denotes the progress-conditioned future representation.
We then refine these progress-conditioned future tokens and predict token-wise
relevance:
\begin{equation}
\begin{gathered}
R_t^l
=
\bar{H}_v^l
+
\operatorname{SelfAttn}
\left(
\bar{H}_v^l
\right), \\
\rho_{t,f}^{l}
=
\operatorname{MLP}_{\omega}
\left(
R_{t,f}^{l}
\right),
\qquad
f=1,\ldots,F .
\end{gathered}
\label{eq:hpim_local_scores}
\end{equation}
where \(R_t^l\) is the refined progress-conditioned representation, and
\(\rho_{t,f}^{l}\) is the relevance logit of the \(f\)-th future latent. The
local relevance weights are normalized over future latents:
\begin{equation}
\omega_{t,f}^{l}
=
F\cdot
\operatorname{Softmax}_{f}
\left(
\rho_{t,f}^{l}
\right),
\qquad
\omega_t^l\in\mathbb{R}_{+}^{F}.
\label{eq:hpim_local_weights}
\end{equation}
The factor \(F\) keeps the average local weight equal to one, so
\(\omega_t^l\) redistributes importance among future latents without introducing
an uncontrolled global scale.


\vspace{2pt}
\noindent\textbf{Progress-Conditioned Attention Modulation.}
We combine the global gate, local future-latent relevance, and layer-wise
schedule into the final attention coefficient. Since the effect of
progress-conditioned modulation may vary across MoT depths, we use a
middle-emphasized layer coefficient:
\begin{equation}
\eta_l
=
\delta
+
(\psi-\delta)
\left[
\sin
\left(
\frac{\pi l}{N_{\mathrm{blk}}-1}
\right)
\right]^{\nu},
\qquad
l=0,\ldots,N_{\mathrm{blk}}-1 .
\label{eq:hpim_layer_schedule}
\end{equation}
Here, \(N_{\mathrm{blk}}\) is the number of MoT layers, \(\psi\) and \(\delta\)
are the maximum and minimum modulation strengths, and \(\nu>0\) controls the
sharpness of the middle-layer emphasis. This schedule assigns weaker modulation
to very shallow and very deep layers, while allowing stronger modulation in
intermediate layers.
The final progress-conditioned coefficient for the \(f\)-th future latent at
layer \(l\) is:
\begin{equation}
\beta_{t,f}^{l}
=
\eta_l
\left[
(1-d_t)
+
d_t \omega_{t,f}^{l}
\right].
\label{eq:hpim_final_coefficient}
\end{equation}
When \(d_t\) is small, \(\beta_{t,f}^{l}\) becomes close to a uniform layer-wise
scaling, reducing token-specific reliance on imagination. When \(d_t\) is large,
the local relevance \(\omega_{t,f}^{l}\) becomes active and differentiates which
future latents should be emphasized. Thus, \(d_t\) controls whether the current
progress should rely on future imagination, while \(\omega_{t,f}^{l}\) controls
which future latents are most relevant.

HPIM injects \(\beta_{t,f}^{l}\) into the MoT attention path from action-token
queries to future-latent keys. Let \(C_{a\rightarrow v}^{l}(i,f)\) denote the
original attention logit from the \(i\)-th action token to the \(f\)-th future
latent:
\begin{equation}
C_{a\rightarrow v}^{l}(i,f)
=
\frac{
(q_{a,i}^{l})^{\top} k_{v,f}^{l}
}{
\sqrt{d_h}
},
\label{eq:hpim_original_logit}
\end{equation}
where \(q_{a,i}^{l}\) is the action-token query, \(k_{v,f}^{l}\) is the
future-latent key, and \(d_h\) is the attention head dimension. We inject the
progress-conditioned coefficient as an additive log-bias:
\begin{equation}
\widetilde{C}_{a\rightarrow v}^{l}(i,f)
=
C_{a\rightarrow v}^{l}(i,f)
+
\log
\left(
\beta_{t,f}^{l}
+
\epsilon
\right),
\label{eq:hpim_logit_modulation}
\end{equation}
where \(\epsilon\) is a small constant for numerical stability.

To preserve the original MoT attention normalization, let
\(\mathcal{K}_{i}^{l}\) denote the complete set of valid keys accessible to the
\(i\)-th action-token query, and let
\(\mathcal{K}_{\mathrm{fut}}^{l}\subseteq\mathcal{K}_{i}^{l}\) denote the subset
of future-latent keys. We use \(\kappa_l(f)\in
\mathcal{K}_{\mathrm{fut}}^{l}\) to denote the position of the \(f\)-th future
latent in the complete key sequence. The complete modulated attention logits are:
\begin{equation}
\widetilde{C}_{a}^{l}(i,k)
=
\begin{cases}
\widetilde{C}_{a\rightarrow v}^{l}(i,f),
&
k=\kappa_l(f),\quad f\in\{1,\ldots,F\},
\\[1mm]
C_{a}^{l}(i,k),
&
k\notin\mathcal{K}_{\mathrm{fut}}^{l},
\end{cases}
\label{eq:hpim_complete_logit}
\end{equation}
where \(C_a^{l}(i,k)\) denotes the original attention from the \(i\)-th
action-token query to the \(k\)-th key. The modulated attention weight assigned
to the \(f\)-th future latent is:
\begin{equation}
\widetilde{A}_{a\rightarrow v}^{l}(i,f)
=
\frac{
\exp
\left(
\widetilde{C}_{a\rightarrow v}^{l}(i,f)
\right)
}{
\displaystyle
\sum_{k\in\mathcal{K}_{i}^{l}}
\exp
\left(
\widetilde{C}_{a}^{l}(i,k)
\right)
}.
\label{eq:hpim_attention_weight}
\end{equation}
Only the action-to-future attention path is modified; all other MoT attention
interactions remain unchanged. This preserves the WAM
backbone while making the influence of imagined futures adaptive to execution
progress.

\vspace{2pt}
\noindent\textbf{Training Objectives for Hierarchical Modulation.}
The two components of HPIM are trained with different signals:

\noindent\textbf{1) Inter-progress global gate.}
For the inter-progress global gate \(d_t\), we automatically construct
frame-level weak labels from demonstrations through a keyframe annotation
pipeline, avoiding dense manual progress annotations whose boundaries are often
subjective in long-horizon manipulation. The pipeline detects coarse interaction
anchors from robot execution traces, such as gripper-state transitions,
contact-related changes, end-effector motion patterns, and task-specific
low-dimensional cues. These anchors indicate central moments of contact-rich
manipulation, including grasping, releasing, insertion, and placement.

Following the finding in
Sec.~\ref{sec:rethinking_imagination_usage}, we assign weak labels according to
the expected usefulness of imagined futures. Timesteps around detected
interaction anchors are labeled as interaction/contact/place phases with
\(y_t^d=0\), where future imagination should be used more conservatively.
Timesteps outside these interaction windows are labeled as
transit/search/approach phases with \(y_t^d=1\), where imagined futures are
expected to provide useful look-ahead guidance. Frames or trajectories with
unreliable keyframe cues are excluded by a validity mask.

Since \(d_t\in[0,1]\) is the predicted inter-progress reliance defined in
Eq.~\ref{eq:hpim_global_gate}, we supervise it with a cross-entropy loss:
\begin{equation}
\mathcal{L}_{\mathrm{gate}}
=
-\frac{1}{|\mathcal{V}_d|}
\sum_{t\in\mathcal{V}_d}
\left[
y_t^d \log d_t
+
(1-y_t^d)\log(1-d_t)
\right],
\label{eq:hpim_gate_loss}
\end{equation}
where \(\mathcal{V}_d\) denotes the timesteps with reliable weak labels. This
loss encourages the global gate to increase future reliance during transit-like
stages and to reduce future reliance during contact-rich interaction.

\noindent\textbf{2) Intra-progress future-latent relevance.}
For the intra-progress future-latent relevance \(\omega_{t,f}^{l}\), we do not
use token-level supervision. Such labels are difficult to define because the
usefulness of each imagined future latent depends on both the current progress
state and the downstream action-generation objective. Instead,
\(\omega_{t,f}^{l}\) is optimized end-to-end through the joint WAM objective, so
that future latents receiving higher relevance are those that better support
action prediction under the current progress condition. In this way, the local
relevance branch is not constrained by hand-crafted token labels, but is learned
toward improving progress-conditioned action generation.
The overall training objective for HPIM is defined as follows:
\begin{equation}
\mathcal{L}_{\mathrm{HPIM}}
=
\mathcal{L}_{\mathrm{act}}
+
\mathcal{L}_{\mathrm{vid}}
+
\lambda_{\mathrm{gate}}
\mathcal{L}_{\mathrm{gate}},
\label{eq:hpim_training_objective}
\end{equation}
where \(\mathcal{L}_{\mathrm{act}}\) and \(\mathcal{L}_{\mathrm{vid}}\) are the
standard action and future-latent generation losses of the joint WAM, and
\(\lambda_{\mathrm{gate}}\) balances the weak supervision for the
inter-progress global gate.

\subsection{Training and Inference Procedure}
\label{sec:training_inference}

\noindent\textbf{Stage-wise Training.}
ProWAM is trained in two stages. In the first stage, we train SS-DTPE
independently on demonstration trajectories to learn online progress
representations. 
This stage is optimized with the self-supervised objective
\(\mathcal{L}_{\mathrm{SS\mbox{-}DTPE}}\) defined in
Eq.~\ref{eq:ss_dtpe_loss}. After training, SS-DTPE is frozen and used as
a progress provider for the joint WAM.

In the second stage, the frozen SS-DTPE provides progress tokens
\(Z_t^{\mathrm{prog}}\), and HPIM uses them to modulate the action-to-future
attention path inside the MoT layers. The joint WAM and HPIM are then trained
with the objective \(\mathcal{L}_{\mathrm{HPIM}}\) defined in
Eq.~\ref{eq:hpim_training_objective}. This stage enables the action generator
to learn how to use future latents under progress-conditioned attention
modulation, while keeping progress estimation decoupled from action learning.

\noindent\textbf{Online Inference.}
During inference, ProWAM does not require offline progress annotations or
precomputed progress caches. At the beginning of each episode, the progress
memory of SS-DTPE is reset, and a start-action token is used when no previous
action is available. At each replanning step, SS-DTPE updates
\(Z_t^{\mathrm{prog}}\) from the current observation, the previously executed
action, the instruction, and the previous progress state. The updated progress
tokens are then passed to HPIM, which modulates the action-to-future attention
inside the joint WAM. The resulting action chunk is executed in a receding-horizon
manner, and the newly executed action is used for the next progress update.

This online procedure allows ProWAM to continuously refresh its progress state
and adapt future-latent utilization as execution evolves, without changing the
standard action-chunk rollout interface of the WAM backbone.

\section{Experiments}

\begin{figure*}[t]
    \centering
    \includegraphics[width=\textwidth]{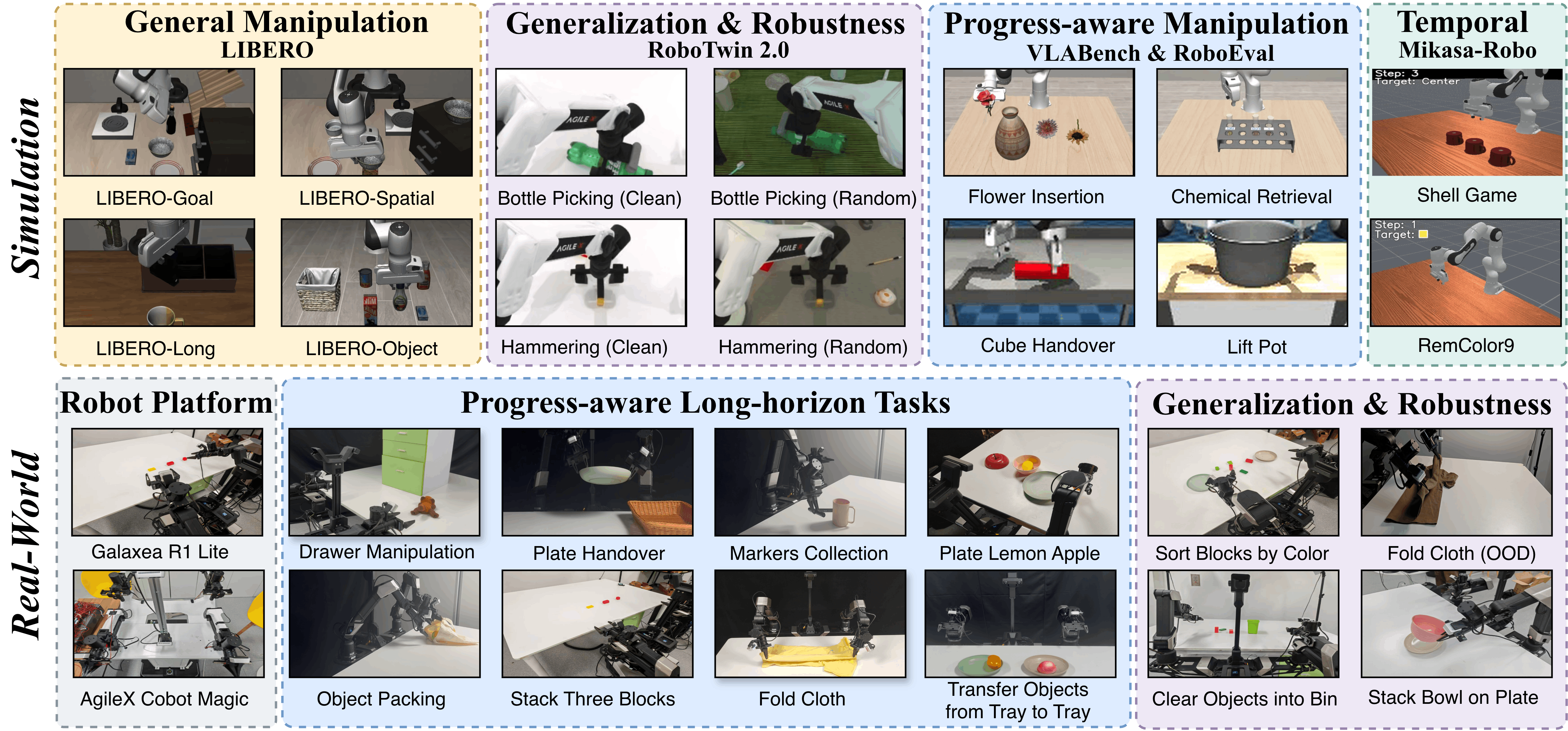}
    \vspace{-14pt}
\caption{\textbf{Experimental Setup.}
\textbf{Top:} Simulation evaluation on LIBERO, RoboTwin2.0, VLABench, RoboEval,
and Mikasa-Robo, covering general manipulation, robustness, progress-aware
execution, and temporal memory.
\textbf{Bottom:} Real-world evaluation on the Galaxea R1 Lite and AgileX Cobot
Magic platforms, covering progress-aware long-horizon manipulation and
robustness to object and scene variations.}
    \label{fig:experimental_setup}
    \vspace{-12pt}
\end{figure*}
\subsection{Implementation Details}
\label{sec:implementation_details}

\noindent\textbf{Model Details.}
Following Fast-WAM~\cite{yuan2026fast}, we use the pretrained Wan2.2-5B model~\cite{wan2025wan}, including the
video DiT, text encoder, and video VAE, while the action branch is implemented as
a 1B action expert with hidden dimension \(d_a=1024\). The action horizon is set to \(H_a=32\).
Video frames are temporally downsampled by \(4\times\), resulting in 9 video
frames per chunk.
For SS-DTPE, the short-term action-observation encoder uses hidden dimension
256, 8 attention heads, and an FFN expansion ratio of 4. The instruction is
represented by 4096-dimensional text features, which condition the visual tokens.
The visual tokenizer uses channel dimension 256 and a spatial downsampling factor
of 16, and the number of action tokens is determined automatically from the
action dimension. The long-term progress aggregator maintains 8 progress tokens
and uses a Mamba-based path with state dimension 16, convolution width 4, and
expansion factor 2.
For the progress objectives, we set the semantic-prefix temperature to
\(\tau_p=0.1\), and the
temporal contrastive temperature to \(\tau_c=0.1\). The semantic alignment and
temporal structure terms use unit weights, and the monotonic order weight is
0.2. For HPIM, we use the  layer schedule with
\(\psi=1.0\), \(\delta=0.1\), and \(\nu=1.0\). The inter-progress gate loss
weight is \(\lambda_{\mathrm{gate}}=0.05\), and the attention log-bias 
constant is \(\epsilon=10^{-4}\).

\vspace{2pt}
\noindent\textbf{Training Details.}
ProWAM is trained in two stages. In Stage 1, SS-DTPE is trained independently
for 5 epochs on 4 NVIDIA H100 GPUs with a per-GPU batch size of 4. In Stage 2,
the trained SS-DTPE is frozen, and the WAM with HPIM is trained for 10
epochs on 8 NVIDIA H100 GPUs with a per-GPU batch size of 16. Both stages use
AdamW with learning rate \(1\times10^{-4}\), weight decay \(1\times10^{-2}\), a
cosine learning-rate schedule, and gradient clipping
at 1.0. 

\vspace{-4pt}
\subsection{Simulation and Real-World Setup}
\label{sec:simulation_real_world_setup}
Fig.~\ref{fig:experimental_setup} provides an overview of the
simulation benchmarks, robotic platforms, and real-world evaluation tasks:

\vspace{2pt}
\noindent\textbf{Simulation Benchmarks.}
We evaluate ProWAM on five simulation benchmarks from four complementary
perspectives: 1) General Manipulation on \textbf{LIBERO}~\cite{liu2023libero};
(2) Robustness and Generalization on 
\textbf{RoboTwin2.0}~\cite{chen2025robotwin};
(3) Progress-aware Manipulation on \textbf{VLABench}~\cite{zhang2025vlabench} and \textbf{RoboEval}~\cite{wang2025roboeval}; and
(4) Temporal and Memory-dependent Manipulation on \textbf{Mikasa-Robo}~\cite{cherepanov2025memory}.
For fair comparison, all methods are evaluated under the standard settings. 

\vspace{2pt}
\noindent\textbf{Real-World Setup.}
We evaluate ProWAM on the Galaxea R1 Lite~\cite{galaxea} and AgileX Cobot
Magic~\cite{aloha} platforms. The Galaxea R1 Lite is a 23-DoF mobile bimanual robot with a wheeled base, articulated torso, dual 6-DoF arms, and multi-view cameras for global and wrist-level observations.
The AgileX Cobot Magic, built upon Mobile ALOHA, features a mobile base, dual 6-DoF arms, parallel grippers, and multi-view sensing. 
Our real-world evaluation covers two complementary categories:
\textbf{progress-aware long-horizon manipulation} and
\textbf{generalization and robustness}. The long-horizon tasks consist of
multiple execution stages and phase transitions, requiring the policy to track
task progress online and maintain coherent advancement throughout the
trajectory. They include Drawer Manipulation, Plate Handover, Markers
Collection, Plate Lemon Apple, Object Packing, Stack Three Blocks, Fold
Cloth, and Transfer Objects from Tray to Tray. The generalization and robustness
tasks include Sort Blocks by Color, Clear
Objects into Bin, Fold Cloth under out-of-distribution conditions, and Stack
Bowl on Plate.

For each task, we collect 150--200 expert demonstrations and conduct 50
evaluation rollouts. Generalization and robustness are evaluated under varying
lighting conditions and scene configurations, while object positions are
randomized in all trials to avoid overfitting to fixed layouts.

\vspace{2pt}
\noindent\textbf{Baselines.}
We compare ProWAM with representative state-of-the-art methods from three
categories: (1)~\textbf{General VLA Models}, such as
OpenVLA-OFT~\cite{kim2025fine}, \(\pi_0\)~\cite{black2024pi_0},
\(\pi_{0.5}\)~\cite{intelligence2025pi_}, PD-VLA~\cite{song2025pd},
SmolVLA~\cite{shukor2025smolvla}, and
MemoryVLA~\cite{shi2025memoryvla};
(2)~\textbf{Predictive-based VLA Models}, such as
CoT-VLA~\cite{zhao2025cot}, DreamVLA~\cite{zhang2026dreamvla},
and
\(\mathcal{F}_1\)~\cite{lv2025f1}; and
(3)~\textbf{World Action Models}, such as
Motus~\cite{bi2026motus}, LingBot-VA~\cite{li2026causal},
WorldVLA~\cite{cen2025worldvla}, 
Fast-WAM~\cite{yuan2026fast}, and
Light-WAM~\cite{li2026light}.

\vspace{-5pt}
\subsection{Evaluation on Simulation Benchmarks}
\label{sec:Evaluation on Simulation Benchmarks}
\begin{table}[t]
   \caption{\textbf{General Robotic Manipulation Performance on LIBERO.}
Comparison of task success rates across the four LIBERO suites. The average is
computed over the four suites, and the best result
is highlighted in bold.}
    \label{tab:libero_main}
    \vspace{-4pt}
    \centering
    \footnotesize
    \setlength{\tabcolsep}{6pt}
    \begin{tabular}{lccccc}
        \hline
        \textbf{Method} & \textbf{Spatial} & \textbf{Object} & \textbf{Goal} & \textbf{Long} & \textbf{Avg.} \\
        \hline

        \rowcolor{gray!20}
        \multicolumn{6}{c}{\textit{General VLA Models}} \\
        \hline
        OpenVLA-OFT~\cite{kim2025fine} &97.6  &98.4     &97.9  &94.5  & 97.1 \\
        \(\pi_0\)~\cite{black2024pi_0} &96.8 &98.8 &95.8 &85.2 &94.2   \\
        \(\pi_{0.5}\)~\cite{intelligence2025pi_} &98.8& 98.2& 98.0& 92.4& 96.9  \\ 
        PD-VLA~\cite{song2025pd} &95.5&96.7  &94.9  &91.7    &94.7  \\
        SmolVLA~\cite{shukor2025smolvla} &93.0 &94.0 &91.0 &77.0 &88.8 \\
        MemoryVLA~\cite{shi2025memoryvla} &98.4 &98.4 &96.4 &93.4 &96.7  \\
        \hline

        \rowcolor{gray!20}
        \multicolumn{6}{c}{\textit{Predictive-based VLA Models}} \\
        \hline
        Seer~\cite{tian2025predictive}&-  & - & - &87.7  &-  \\
        UniVLA~\cite{bu2025univla} &96.5  &96.8  & 95.6 & 92.0 & 95.2 \\
        CoT-VLA~\cite{zhao2025cot} &87.5  &91.6  & 87.6 &69.0&83.9  \\
        DreamVLA~\cite{zhang2026dreamvla} & 97.5 & 94.0 & 89.5 & 89.5 & 92.6 \\
        \(\mathcal{F}_1\)~\cite{lv2025f1} &98.2  & 97.8 & 95.4  & 91.3  & 95.7 \\
        \hline

        \rowcolor{gray!20}
        \multicolumn{6}{c}{\textit{World Action Models}} \\
        \hline
        Motus~\cite{bi2026motus} &96.8 &99.8 &96.6 &97.6 &97.7  \\
        LingBot-VA~\cite{li2026causal} &98.5 &99.6 &97.2 &\textbf{98.5 }&98.5  \\
        WorldVLA~\cite{cen2025worldvla} &87.6 &96.2 &83.4 &60.0 &81.8 \\
        VLA-JEPA~\cite{sun2026vla} & 96.2 & 99.6 & 97.2 & 95.8 & 97.2  \\
        Fast-WAM~\cite{yuan2026fast} &  98.2 &\textbf{100.0} &97.0 &95.2 &97.6  \\
        ImageWAM~\cite{zhang2026imagewam} & 97.2 &99.2 &\textbf{98.8} &98.4 &98.4 \\
        Light-WAM~\cite{li2026light} &98.2 &99.6 &97.8 &93.0 &97.2  \\
        MaskWAM~\cite{yu2026maskwam} & 98.8 &\textbf{100.0} &98.2 &96.4 &98.4  \\
        \hline

        \rowcolor[HTML]{ECDFF2}
        \textbf{ProWAM (Ours)} & \textbf{99.6 } &  \textbf{100.0} & \textbf{ 98.8} & 97.8 & \textbf{99.1 } \\
        \hline
    \end{tabular}
    \vspace{-5pt}
\end{table}

\noindent\textbf{Results on LIBERO.}
Table~\ref{tab:libero_main} presents the results across the four LIBERO suites.
ProWAM achieves an average success rate of \(99.1\%\), outperforming all
compared methods. Compared with OpenVLA-OFT~\cite{kim2025fine},
\(\mathcal{F}_1\)~\cite{lv2025f1}, and the best-performing competing WAM,
ProWAM improves the average success rate by \(2.0\), \(3.4\), and \(0.6\)
percentage points, respectively. Notably, ProWAM surpasses
Fast-WAM~\cite{yuan2026fast} by \(1.5\) percentage points on average. Unlike
Fast-WAM, which does not explicitly regulate future-latent utilization according
to execution progress, ProWAM adaptively modulates the influence of imagined
futures throughout task execution. It achieves the best or tied-best results on
Spatial (\(99.6\%\)), Object (\(100.0\%\)), and Goal (\(98.8\%\)), while
maintaining competitive performance on Long (\(97.8\%\)). 

\begin{table}[t]
    \caption{\textbf{Robustness and Generalization Performance on RoboTwin2.0.}
    Following the standard protocol~\cite{li2026causal,bi2026motus,yuan2026fast},
    we report success rates under clean and randomized settings.
    Clean uses fixed initial configurations, while Randomized varies object
    poses and scene layouts. \textbf{Embodied PT.}\ indicates whether embodied
    pre-training is used.}
    \label{tab:robotwin_main}
    \vspace{-4pt}
    \centering
    \footnotesize
    \setlength{\tabcolsep}{3pt}
    \begin{tabular}{lcccc}
        \hline
        \textbf{Method}
        & \textbf{Embodied PT.}
        & \textbf{Clean}
        & \textbf{Randomized}
        & \textbf{Avg.} \\
        \hline

        \rowcolor{gray!20}
        \multicolumn{5}{c}{\textit{General VLA Models}} \\
        \hline
        \(\pi_0\)~\cite{black2024pi_0}
        & \(\checkmark\) & 65.9 & 58.4 & 62.2 \\
        \(\pi_{0.5}\)~\cite{intelligence2025pi_}
        & \(\checkmark\) & 82.7 & 76.8 & 79.8 \\
        X-VLA~\cite{zheng2025x}
        & \(\checkmark\) & 72.9 & 72.8 & 72.9 \\
        ABot-M0~\cite{yang2026abot} & \(\checkmark\)&  86.1 & 85.1 & 85.6 \\
        \hline

        \rowcolor{gray!20}
        \multicolumn{5}{c}{\textit{World Action Models}} \\
        \hline
        Motus~\cite{bi2026motus}
        & \(\checkmark\) & 88.7 & 87.0 & 87.8 \\
        LingBot-VA~\cite{li2026causal}
        & \(\checkmark\) & 92.9 & 91.5 & 92.2 \\
        Fast-WAM~\cite{yuan2026fast}
        & \(\times\) & 91.9 & 91.8 & 91.9 \\
        Light-WAM~\cite{li2026light}
        & \(\times\) & 76.4 & 76.3 & 76.4 \\
        Being-H0.7~\cite{luo2026being}
        & \(\checkmark\) & 90.2 & 89.6 & 89.9 \\
        GigaWorld-Policy~\cite{ye2026gigaworld}
        & \(\checkmark\) & 86.4 & 85.0 & 85.7 \\
        X-WAM~\cite{guo2026unified}
        & \(\checkmark\) & 89.8 & 90.7 & 90.2 \\
        FlowWAM~\cite{flowwam}
        & \(\checkmark\) & 92.9 & 92.1 & 92.5 \\
        \hline
        \rowcolor[HTML]{ECDFF2}
        \textbf{ProWAM (Ours)}
        & \(\times\) & \textbf{93.9} & \textbf{92.8} & \textbf{93.4} \\
        \hline
    \end{tabular}
    \vspace{-8pt}
\end{table}

\begin{table}[t]
    \caption{\textbf{Progress-Aware Manipulation Performance on VLABench.}
    We report Success Rate (SR), Intention Score (IS), and Progress Score (PS).
    \(^{*}\) denotes results reproduced by us under the same
    protocol.}
    \label{tab:vlabench_main}
    \vspace{-4pt}
    \centering
    \footnotesize
    \setlength{\tabcolsep}{16pt}
    \begin{tabular}{lccc}
        \hline
        \textbf{Method}
        & \textbf{SR} \(\uparrow\)
        & \textbf{IS} \(\uparrow\)
        & \textbf{PS} \(\uparrow\) \\
       
        \hline
        \(\pi_0\)~\cite{black2024pi_0}
        & 47.0 & 67.8 & 62.7 \\
        \(\pi_0\)-FAST~\cite{pertsch2025fast}
        & 56.2 & 72.4 & 66.8 \\
        \(\pi_{0.5}\)~\cite{intelligence2025pi_}
        & 65.4 & 80.4 & 77.8 \\
        X-VLA~\cite{zheng2025x}
        & -- & -- & 67.8 \\
    
        ACoT-VLA~\cite{zhong2026acot}
        & -- & 79.8 & 66.1 \\
        Motus\(^{*}\)~\cite{bi2026motus}
        & 54.4& 78.6 & 70.2\\
        Fast-WAM\(^{*}\)~\cite{yuan2026fast}
        & 58.4 & 81.2 & 74.5 \\
        \hline
        \rowcolor[HTML]{ECDFF2}
        \textbf{ProWAM (Ours)}
        & \textbf{68.4} & \textbf{85.4} & \textbf{81.2} \\
        \hline
    \end{tabular}
    \vspace{-5pt}
\end{table}

\vspace{2pt}
\noindent\textbf{Results on RoboTwin2.0.}
As shown in Table~\ref{tab:robotwin_main}, ProWAM achieves the best overall
performance on RoboTwin~2.0, reaching \(93.9\%\) and \(92.8\%\) success rates
under the Clean and Randomized settings, respectively, with an average of
\(93.4\%\). It outperforms FlowWAM~\cite{flowwam} by \(1.0/0.7/0.9\) percentage
points on Clean/Randomized/Avg., and improves over Fast-WAM~\cite{yuan2026fast}
by \(1.5\) points on average despite using no additional embodied pre-training.
ProWAM also surpasses all compared methods with embodied pre-training.

These gains are particularly relevant to RoboTwin~2.0, where bimanual
manipulation requires consistent adaptation across transit, approach, and
contact-rich interaction stages under substantial scene variations. By
conditioning imagination utilization on recurrent execution progress, ProWAM
maintains coherent stage transitions while adapting the contribution of future
information to changing control demands. Its consistent gains under both Clean
and Randomized settings demonstrate strong   performance
and robustness to object-pose and scene-layout variations.

\vspace{2pt}
\noindent\textbf{Results on VLABench.}
As shown in Table~\ref{tab:vlabench_main}, ProWAM achieves the best performance
across all three metrics, reaching \(68.4\%\) SR, \(85.4\%\) IS, and
\(81.2\%\) PS. Compared with \(\pi_{0.5}\)~\cite{intelligence2025pi_}, ProWAM improves SR, IS, and PS by
\(3.0\), \(5.0\), and \(3.4\) percentage points, respectively. It also
outperforms Fast-WAM~\cite{yuan2026fast} by \(10.0\) percentage points in SR,
\(4.2\) in IS, and \(6.7\) in PS, demonstrating the benefit of
explicitly conditioning future-latent utilization on execution progress.
The gain in PS is particularly relevant to our progress-aware design, since PS
measures intermediate task completion rather than only terminal success.
Meanwhile, the improvements in IS and SR indicate that ProWAM preserves accurate
instruction grounding and translates this understanding into more reliable task
completion. These consistent gains suggest that recurrent progress modeling
helps the policy identify its current execution stage, while
progress-conditioned imagination modulation enables it to selectively exploit
future information throughout multi-stage manipulation.

\begin{table}[t]
    \caption{\textbf{Progress-Aware Manipulation Performance on RoboEval.}
    We report the mean Success Rate (SR) and Task Progression (TP) across eight
    tasks and 30 task variations.}
    \label{tab:roboeval_main}
    \vspace{-4pt}
    \centering
    \footnotesize
    \setlength{\tabcolsep}{22pt}
    \begin{tabular}{lcc}
        \hline
        \textbf{Method}
        & \textbf{SR} \(\uparrow\)
        & \textbf{TP} \(\uparrow\) \\
        \hline

        ACT~\cite{zhao2023learning}
        & 0.278
        & 0.454 \\

        \(\pi_{0.5}\) (LoRA)~\cite{intelligence2025pi_}
        & 0.105
        & 0.336 \\

        Diffusion Policy~\cite{chi2025diffusion}
        & 0.030
        & 0.200 \\

        X-VLA~\cite{zheng2025x}
        & 0.015
        & 0.114 \\

        GR00T~\cite{bjorck2025gr00t}
        & 0.013
        & 0.136 \\
          Fast-WAM\(^{*}\)~\cite{yuan2026fast}
        & 0.313 & 0.404 \\
        \hline

        \rowcolor[HTML]{ECDFF2}
        \textbf{ProWAM (Ours)}
        & \textbf{0.375}
        & \textbf{0.494} \\
        \hline
    \end{tabular}
    \vspace{-5pt}
\end{table}
\begin{table}[t]
    \caption{\textbf{Temporal and Memory-Dependent Manipulation Performance on
    Mikasa-Robo.}  Following the standard protocol~\cite{shi2025memoryvla, shi2026memoryvla++}, 
    we report success rates (\%) over 100 evaluation episodes for each task.
    SGT, IM, and RC3/5/9 denote ShellGameTouch, InterceptMedium, and
    RememberColor with 3/5/9 candidate objects, respectively. }
    \label{tab:mikasa_main}
    \vspace{-4pt}
    \centering
    \footnotesize
    \setlength{\tabcolsep}{6pt}
    \begin{tabular}{lcccccc}
        \hline
        \textbf{Method}
        & \textbf{SGT}
        & \textbf{IM}
        & \textbf{RC3}
        & \textbf{RC5}
        & \textbf{RC9}
        & \textbf{Avg.} \\
        \hline
        DP~\cite{chi2025diffusion}
        & 23 & -- & 3 & -- & -- & -- \\
        PTP~\cite{villasevil2025learning}
        & 26 & -- & 4 & -- & -- & -- \\
        MaIL~\cite{jia2024mail}
        & 27 & -- & 11 & -- & -- & -- \\
        Octo~\cite{team2024octo}
        & 46 & 39 & 45 & 17 & 11 & 31.6 \\
        CronusVLA~\cite{li2025cronusvla}
        & 32 & 5 & 31 & 13 & 9 & 18.0 \\
        SpatialVLA~\cite{qu2025spatialvla}
        & 23 & 27 & 27 & 17 & 11 & 21.0 \\
        OpenVLA-OFT~\cite{kim2025fine}
        & 47 & 14 & \textbf{59} & 16 & 6 & 28.4 \\
        \(\pi_0\)~\cite{black2024pi_0}
        & 33 & 42 & 35 & 22 & 15 & 29.4 \\
        MemoryVLA~\cite{shi2025memoryvla}
        & 88 & 24 & 44 & \textbf{30} & \textbf{20} & 41.2 \\
        MemoryVLA++~\cite{shi2026memoryvla++}
        & \textbf{97} & 40 & 50 & 19 & 16 & 44.4 \\
        \hline

        \rowcolor[HTML]{ECDFF2}
        \textbf{ProWAM (Ours)}
        & 94
        & \textbf{68}
        & 48
        & 24
        & \textbf{20}
        & \textbf{50.8} \\
        \hline
    \end{tabular}
    \vspace{-10pt}
\end{table}

\vspace{2pt}
\noindent\textbf{Results on RoboEval.}
As shown in Table~\ref{tab:roboeval_main}, ProWAM achieves an SR of \(0.375\)
and a TP of \(0.494\), outperforming all methods included in the comparison.
Compared with our reproduced Fast-WAM baseline, ProWAM improves SR and TP by
\(0.062\) and \(0.090\), respectively. It also surpasses ACT~\cite{zhao2023learning} by \(0.097\) in
SR and \(0.040\) in TP.
The improvement in TP is particularly relevant to our progress-aware design,
as TP measures the intermediate stages completed by the policy rather than only
terminal task success. The larger gain over Fast-WAM~\cite{yuan2026fast} in TP suggests that
explicit progress modeling enables the policy to maintain more consistent
advancement through multi-stage execution. By
accumulating action-observation evidence into a recurrent progress state and
conditioning future-latent utilization on the inferred execution stage, ProWAM
can better distinguish completed and remaining sub-stages and exploit predictive
information according to the current control demand.

\vspace{2pt}
\noindent\textbf{Results on Mikasa-Robo.}
As shown in Table~\ref{tab:mikasa_main}, ProWAM achieves the highest average
success rate of \(50.8\%\) across five temporal and memory-dependent tasks,
outperforming MemoryVLA++~\cite{shi2026memoryvla++} and
MemoryVLA~\cite{shi2025memoryvla} by \(6.4\) and \(9.6\) percentage points,
respectively. It achieves the best result on InterceptMedium (\(68\%\)),
exceeding MemoryVLA++ by \(28\) points, and ties the best performance on
RememberColor9 (\(20\%\)), while reaching \(94\%\) on ShellGameTouch.

These results highlight the benefit of recurrent progress modeling for tasks
requiring temporal memory and state disambiguation. In particular, the strong
gain on InterceptMedium suggests that Mamba-based progress aggregation helps
preserve motion history, while progress-conditioned imagination emphasizes
future cues relevant to the current interception stage. Together, they enable
ProWAM to integrate accumulated evidence with anticipated dynamics for
memory-dependent control.
\vspace{-5pt}
\subsection{Evaluation on Real-World Tasks}
\label{sec:Evaluation on Real-World}
\begin{table*}[t]
   \caption{\textbf{Real-World Progress-Aware Long-Horizon Manipulation Results.}
    \(^{\dagger}\) denotes results reproduced under the same training and
deployment settings as ProWAM. Each entry reports the cumulative
stage-completion success rate (\%) over 50 evaluation trials. Avg.\ denotes the
average full-task success rate across all eight tasks, computed from the
completion rate of the final stage. The best result in each column is
highlighted in bold.}
    \label{tab:real_world_long_horizon}
    \vspace{-7pt}
    \centering
    \footnotesize
    \setlength{\tabcolsep}{1pt}
    \renewcommand{\arraystretch}{1.05}

    \resizebox{\textwidth}{!}{
    \begin{tabular}{l|ccc|ccc|ccc|cc|cc|ccc|ccc|ccc|c}
        \toprule
        \textbf{Method}
        & \multicolumn{3}{c|}{\textbf{Drawer M.}}
        & \multicolumn{3}{c|}{\textbf{Fold C.}}
        & \multicolumn{3}{c|}{\textbf{Plate L.}}
        & \multicolumn{2}{c|}{\textbf{Object P.}}
        & \multicolumn{2}{c|}{\textbf{Marker C.}}
        & \multicolumn{3}{c|}{\textbf{Stack T.}}
        & \multicolumn{3}{c|}{\textbf{Tray T.}}
        & \multicolumn{3}{c|}{\textbf{Plate H.}}
        & \textbf{Avg.} \\

        & \textbf{Pull}
        & \textbf{+Place}
        & \textbf{+Push}

        & \textbf{Grasp}
        & \textbf{+Fold}
        & \textbf{+Release}

        & \textbf{Open}
        & \textbf{+Retrieve}
        & \textbf{+Cover}

        & \textbf{Open}
        & \textbf{+Pack}

        & \textbf{Insert}
        & \textbf{+Insert}

        & \textbf{Grasp}
        & \textbf{+Stack}
        & \textbf{+Stack}

        & \textbf{Pick}
        & \textbf{+Transfer}
        & \textbf{+Place}

        & \textbf{Pick}
        & \textbf{+Pass}
        & \textbf{+Place}

        & \textbf{(\%)} \\
        \hline

        \rowcolor[HTML]{EFEFEF}
        \multicolumn{24}{c}{\textit{\footnotesize Galaxea R1 Lite platform}} \\
        \hline
        OpenVLA\(^{\dagger}\)
         & 60 & 52 & 40
        & 40  &34  &20 
        & 56 & 46 & 38
        & 48 & 46
        &60 &54  &54 
        &52 &48  &40 
        &34 &20   
        &56 &54  &50 &40
        \\
        OFT\(^{\dagger}\)
        & 70 & 68 & 60
        & 48 & 40 & 28
        & 50 & 42 & 40
        & 60 & 54
        & 66 &58
        &60  &58  &54 
        &44  &42  &36 
        &66 &64  &64 
        &49  \\

        \(\pi_{0.5}\)\(^{\dagger}\)
        & 74 & 74 & 66
        & 68 & \textbf{66} & 58
        & 70 & 64 & 60
        & 74 & 64
        &\textbf{74} &\textbf{68}  &78 
        &74 &70  &64 & 60 &52
        &\textbf{78} &\textbf{76}  &\textbf{74} 
        &64
        \\

        DreamVLA\(^{\dagger}\)
        & 58 & 58 & 54
        & 50 & 40 & 34
        & 54 & 52 & 48
        & 54 & 48
        & 68 & 56
        &56  &50 
       &50&40  & 36
        &34 &46  &44 
        & 40&46\\
        Fast-WAM\(^{\dagger}\)
         &68  &62  &56 
        &68  &60  &52
        &62  &60  &58
        &54  &54 
        &66  &60 
        &70 &66  &66
       &62 &54  &48 
        &64 &60  &56 
        &56  \\
        \hline

        \rowcolor[HTML]{ECDFF2}
        \textbf{ProWAM}
        & \textbf{80} & \textbf{78} & \textbf{76}
        & \textbf{70} & \textbf{66} & \textbf{66}
        & \textbf{76} & \textbf{70} & \textbf{70}
        & \textbf{84} & \textbf{78}
        & 72 & 66
        & \textbf{84} & \textbf{78} & \textbf{78}
        & \textbf{72} & \textbf{68} & \textbf{64}
        & 76 & 74 & \textbf{74}
        & \textbf{72} \\
        \hline

        \rowcolor[HTML]{EFEFEF}
        \multicolumn{24}{c}{\textit{\footnotesize AgileX Cobot Magic platform}} \\
        \hline
    
        OFT\(^{\dagger}\)
        &68  &68 &62 
       &52  &44 &24
        &54  &48 &46 
        &66  &60 
        &64  &64 
        &58  &56 &56 
        &48  &44 &34 
        &70  &64 &62 
        &51 \\

        Fast-WAM\(^{\dagger}\)
         &70  &70 &64 
       &64  &58 &48 
        &66  &62 &56 
        &58  &56 
        &68  &66 
        &72  &64 &64 
        &56  &56 &52 
        &56  &54 &50 
        &57 \\
        \hline

        \rowcolor[HTML]{ECDFF2}
        \textbf{ProWAM}
        & \textbf{82} & \textbf{80} & \textbf{70}
        & \textbf{68} & \textbf{68} & \textbf{66}
        & \textbf{78} & \textbf{74} & \textbf{74}
        & \textbf{82} & \textbf{76}
        & \textbf{76} & \textbf{68}
        & \textbf{88} & \textbf{86} & \textbf{80}
        & \textbf{70} & \textbf{62} & \textbf{62}
        & \textbf{78} & \textbf{74} & \textbf{72}
        & \textbf{71} \\
        \bottomrule
    \end{tabular}
    }

    \vspace{-10pt}
\end{table*}
\noindent\textbf{Results on Progress-Aware Long-Horizon Tasks.}
Table~\ref{tab:real_world_long_horizon} reports cumulative stage-completion
rates and full-task success rates on two real-robot platforms. On the Galaxea
R1 Lite, ProWAM achieves the highest average full-task success rate of
\(72\%\), outperforming \(\pi_{0.5}\)~\cite{intelligence2025pi_},
Fast-WAM~\cite{yuan2026fast}, and DreamVLA~\cite{zhang2026dreamvla} by
\(8\), \(16\), and \(26\) percentage points, respectively. On the AgileX
Cobot Magic, ProWAM reaches \(71\%\), exceeding Fast-WAM and
OpenVLA-OFT~\cite{kim2025fine} by \(14\) and \(20\) percentage points, respectively.
The stage-wise results further show that ProWAM preserves execution progress
more effectively as the task unfolds. On the Galaxea
R1 Lite, its average completion rate
decreases from approximately \(77\%\) at the first stage to \(72\%\) at full
task completion, corresponding to a drop of only \(5\) percentage points. In comparison,
\(\pi_{0.5}\) decreases from \(73\%\) to \(64\%\), while Fast-WAM decreases
from \(64\%\) to \(56\%\). The widening performance margin at later stages indicates that ProWAM not only
executes individual manipulation primitives reliably, but also mitigates error
accumulation throughout multi-stage task execution.

\begin{table}[t]
    \caption{\textbf{Real-World Generalization and Robustness Results.}
    \(^{\dagger}\) denotes results reproduced under the same training and
    deployment settings as ProWAM. We report cumulative stage-completion rates
    (\%) over 50 trials with lighting, scene, and object variations. }
    \label{tab:real_world_generalization}
    \vspace{-5pt}
    \centering
    \footnotesize
    \setlength{\tabcolsep}{0.5pt}

    \resizebox{\columnwidth}{!}{
    \begin{tabular}{l|ccc|cc|ccc|cc|c}
        \toprule
        \textbf{Method}
        & \multicolumn{3}{c|}{\textbf{Sort B.}}
        & \multicolumn{2}{c|}{\textbf{Clear O.}}
        & \multicolumn{3}{c|}{\textbf{Fold (OOD)}}
        & \multicolumn{2}{c|}{\textbf{Stack B.}}
        & \textbf{Avg.} \\

        & \textbf{Pick}
        & \textbf{+Sort}
        & \textbf{+Place}

        & \textbf{Pick}
        & \textbf{+Transfer}

        & \textbf{Grasp}
        & \textbf{+Fold}
        & \textbf{+Release}

        & \textbf{Pick}
        & \textbf{+Place}

        & \textbf{(\%)} \\
        \hline

        \rowcolor[HTML]{EFEFEF}
        \multicolumn{12}{c}{\textit{\footnotesize Galaxea R1 Lite platform}} \\
        \hline

        OFT\(^{\dagger}\)
        & 64 & 60 & 56
        & 58& 56
        & 24 & 16 & 14
        & 66 & 62
        & 47 \\

        \(\pi_{0.5}^{\dagger}\)
        & 74 & 70 &70
        & 68 &66
        & 48 & 44 & 42
        & 80 & 76
        & 64 \\

        DreamVLA\(^{\dagger}\)
        & 54 & 50 & 48
        & 52 &52 
        & 28 & 26 & 20
        & 68 & 56
        & 44 \\

        Fast-WAM\(^{\dagger}\)
        & 68 & 66 & 62
        & 66 & 64
        & 42 & 36 & 32
        & 74 & 72
        & 58 \\
        \hline

        \rowcolor[HTML]{ECDFF2}
        \textbf{ProWAM}
        & \textbf{80} & \textbf{78} & \textbf{72}
        & \textbf{72} & \textbf{68}
        & \textbf{58} & \textbf{56} & \textbf{54}
        & \textbf{86} & \textbf{80}
        & \textbf{69} \\
         \hline
         \rowcolor[HTML]{EFEFEF}
        \multicolumn{12}{c}{\textit{\footnotesize AgileX Cobot Magic platform}} \\
        \hline

        OFT\(^{\dagger}\)
        & 70 & 62 & 58
        & 54 & 54
        & 28 & 24 & 16
        & 64 & 60
        & 47 \\

        Fast-WAM\(^{\dagger}\)
        & 72 & 64 & 60
        & \textbf{68} & \textbf{68}
        & 48 & 42 & 34
        & 72 & 66
        & 57 \\
        \hline

        \rowcolor[HTML]{ECDFF2}
        \textbf{ProWAM}
        & \textbf{82} & \textbf{80} & \textbf{76}
        & \textbf{68} & 64
        & \textbf{62} & \textbf{54} & \textbf{52}
        & \textbf{88} & \textbf{80}
        & \textbf{68} \\
        \bottomrule
    \end{tabular}
    }

    \vspace{-5pt}
\end{table}

\vspace{2pt}
\noindent\textbf{Results on Real-World Generalization and Robustness Tasks.}
Table~\ref{tab:real_world_generalization} reports cumulative stage-completion
rates under variations in lighting, scene configurations, and object positions.
On the Galaxea R1 Lite, ProWAM achieves the highest average full-task success
rate of \(69\%\), outperforming
\(\pi_{0.5}\)~\cite{intelligence2025pi_},
Fast-WAM~\cite{yuan2026fast}, and
OpenVLA-OFT~\cite{kim2025fine} by \(5\), \(11\), and \(22\) percentage
points, respectively. On the AgileX Cobot Magic, ProWAM reaches \(68\%\),
exceeding Fast-WAM and OpenVLA-OFT by \(11\) and \(21\) percentage points,
respectively.
The gains are particularly pronounced on Fold Cloth under
out-of-distribution conditions. On the Galaxea R1 Lite platform, ProWAM improves
the final-stage success rate over \(\pi_{0.5}\) and Fast-WAM by \(12\) and
\(22\) percentage points, respectively. On the AgileX Cobot Magic platform, it
surpasses Fast-WAM by \(18\) percentage points.  

\vspace{-6pt}
\subsection{Ablation Study}
\label{sec:ablation_studies}
We conduct ablation studies on the LIBERO simulation benchmark and the two
real-world task categories introduced above, namely progress-aware long-horizon
tasks (\textbf{Long.}) and generalization and robustness tasks
(\textbf{Gen.}), to systematically examine the effectiveness of the proposed
components and design choices. All real-world ablations are conducted on the
Galaxea R1 Lite platform.

\begin{table}[t]
    \caption{\textbf{Ablation on Model Components.}
   We ablate the proposed SS-DTPE and HPIM to evaluate their contributions on LIBERO and real-world manipulation tasks.}
    \label{tab:component_ablation}
    \vspace{-5pt}
    \centering
    \footnotesize
    \setlength{\tabcolsep}{3.5pt}

    \begin{tabular}{cc|cccc|cc}
        \toprule
        \multicolumn{2}{c|}{\textbf{Method}}
        & \multicolumn{4}{c|}{\textbf{LIBERO}}
        & \multicolumn{2}{c}{\textbf{Real-World}} \\
        \cmidrule(lr){1-2}
        \cmidrule(lr){3-6}
        \cmidrule(lr){7-8}

        \textbf{SS-DTPE}
        & \textbf{HPIM}
        & \textbf{Spatial}
        & \textbf{Object}
        & \textbf{Goal}
        & \textbf{Long}
        & \textbf{Long.}
        & \textbf{Gen.} \\
        \midrule

        \(\times\)
        & \(\times\)
        & 96.6
        & 99.6
        & 97.4
        & 94.8
        & 56
        & 58 \\

        \(\checkmark\)
        & \(\times\)
        & 98.4
        & 99.6
        & 98.0
        & 96.2
        & 66
        & 65 \\

        \(\times\)
        & \(\checkmark\)
        & 98.2
        & 99.4
        & 98.2
        & 96.6
        & 62
        & 64 \\
        \midrule

        \rowcolor[HTML]{ECDFF2}
        \(\checkmark\)
        & \(\checkmark\)
        & \textbf{99.6}
        & \textbf{100.0}
        & \textbf{98.8}
        & \textbf{97.8}
        & \textbf{72}
        & \textbf{69} \\
        \bottomrule
    \end{tabular}

    \vspace{-8pt}
\end{table}

\vspace{2pt}
\noindent\textbf{Ablation Studies on Model Components.}
Table~\ref{tab:component_ablation} evaluates SS-DTPE and HPIM under four
controlled settings: \textbf{(1) Baseline}, without either module;
\textbf{(2) SS-DTPE Only}, where the learned progress representation is directly
fused into the action stream without attention modulation;
\textbf{(3) HPIM Only}, where HPIM is conditioned on a stateless context from
the current observation, previous action, and language instruction; and
\textbf{(4) Full ProWAM}, where SS-DTPE recursively estimates execution progress
and HPIM uses it to modulate action-to-future attention.
Both components independently improve over the baseline. SS-DTPE yields gains
of \(1.8/0.6/1.4\) percentage points on LIBERO Spatial/Goal/Long and \(10/7\) points on
real-world Long./Gen., demonstrating the benefit of temporally grounded progress
modeling. HPIM improves the corresponding LIBERO suites by \(1.6/0.8/1.8\)
percentage points and both real-world settings by \(6\), validating adaptive
future-latent modulation. Full ProWAM performs best in every column, improving
over the baseline by \(3.0/0.4/1.4/3.0\) across the LIBERO suites
and \(16/11\) on real-world Long./Gen. These  results demonstrate the
complementarity of the two modules: SS-DTPE provides structured execution
progress, while HPIM translates it into progress-adaptive utilization.

\begin{table}[t]
  \caption{\textbf{Ablation Studies on Long-Term Progress Aggregation in SS-DTPE.}
We compare different aggregation mechanisms while keeping other
components unchanged.}
    \label{tab:ss_dtpe_ablation}
    \vspace{-5pt}
    \centering
    \footnotesize
    \setlength{\tabcolsep}{4.5pt}

    \begin{tabular}{l|cccc|cc}
        \toprule
        \multirow{2}{*}{\textbf{Aggregator}}
        & \multicolumn{4}{c|}{\textbf{LIBERO}}
        & \multicolumn{2}{c}{\textbf{Real-World}} \\
        \cmidrule(lr){2-5}
        \cmidrule(lr){6-7}

        & \textbf{Spatial}
        & \textbf{Object}
        & \textbf{Goal}
        & \textbf{Long}
        & \textbf{Long.}
        & \textbf{Gen.} \\
        \midrule

        w/o Aggregator
        & 96.2 & 99.2 &97.0 & 95.2
        & 58 & 59 \\

        LSTM~\cite{hochreiter1997long}
        & 97.2 & 99.6 & 98.2 & 96.6
        & 60 & 63 \\

        Transformer~\cite{vaswani2017attention}
        & 97.6 & 99.8 & 97.6 & 96.4
        & 62 & 61 \\
        \midrule

        \rowcolor[HTML]{ECDFF2}
        \textbf{Mamba (Ours)}
        & \textbf{99.6}
        & \textbf{100.0}
        & \textbf{98.8}
        & \textbf{97.8}
        & \textbf{72}
        & \textbf{69} \\
        \bottomrule
    \end{tabular}

    \vspace{-5pt}
\end{table}
\begin{table}[t]
\caption{\textbf{Ablations on Self-Supervised Training Objectives in SS-DTPE.}
Sem.\ Align.\ denotes Semantic Progress Alignment, comprising Properties
\textit{i)} and \textit{ii)} introduced in Sec.~\ref{sec:progress_encoding},
while Temp.\ Struct.\ is Temporal Structure Learning.}
    \label{tab:ss_dtpe_objective_ablation}
    \vspace{-5pt}
    \centering
    \footnotesize
    \setlength{\tabcolsep}{2.2pt}

    \begin{tabular}{ccc|cccc|cc}
        \toprule
        \multicolumn{2}{c}{\textbf{Sem. Align.}}
        & \multirow{2}{*}{\shortstack{\textbf{Temp.}\\\textbf{Struct.}}}
        & \multicolumn{4}{c|}{\textbf{LIBERO}}
        & \multicolumn{2}{c}{\textbf{Real-World}} \\
        \cmidrule(lr){1-2}
        \cmidrule(lr){4-7}
        \cmidrule(lr){8-9}

        \textbf{Prop. \textit{i)}}
        & \textbf{Prop. \textit{ii)}}
        &
        & \textbf{Spatial}
        & \textbf{Object}
        & \textbf{Goal}
        & \textbf{Long}
        & \textbf{Long.}
        & \textbf{Gen.} \\
        \midrule

        \(\times\)
        & \(\times\)
        & \(\times\)
        & 96.0 & 98.6 & 96.6 & 94.4
        & 54 & 57 \\
               \(\times\)
        & \(\times\)
        & \(\checkmark\)
        & 96.4 & 99.0 & 97.4 & 94.6
        & 57 & 58 \\

        \(\checkmark\)
        & \(\times\)
        & \(\times\)
        &96.8  &99.0  & 97.2&95.0 
        &59  & 63 \\

        \(\checkmark\)
        & \(\checkmark\)
        & \(\times\)
        & 97.8 & 99.4 & 98.2 & 97.0
        & 67 & 66 \\

        \midrule

        \rowcolor[HTML]{ECDFF2}
        \(\checkmark\)
        & \(\checkmark\)
        & \(\checkmark\)
        & \textbf{99.6}
        & \textbf{100.0}
        & \textbf{98.8}
        & \textbf{97.8}
        & \textbf{72}
        & \textbf{69} \\
        \bottomrule
    \end{tabular}

    \vspace{-8pt}
\end{table}

\vspace{2pt}
\noindent\textbf{Ablation Studies on Long-Term Progress Aggregation.}
Table~\ref{tab:ss_dtpe_ablation} investigates the effect of different
aggregation mechanisms in the long-term progress branch of SS-DTPE. Removing
the long-term aggregator causes consistent performance degradation, especially
on long-horizon and real-world tasks. Compared with this variant, introducing
LSTM improves LIBERO-Long and real-world Long.\ by \(1.4\) and \(2\) percentage
points, respectively, demonstrating the importance of explicitly maintaining
historical progress information. Replacing the aggregator with a Transformer
further improves most LIBERO suites, but provides limited gains on real-world
generalization, suggesting that standard attention-based aggregation is less
effective at preserving long-term execution states under environmental
variations.

The proposed Mamba-based aggregation achieves the best performance across all
benchmarks.  These gains indicate that the selective
state-space modeling of Mamba is more suitable for progress representation,
allowing SS-DTPE to retain relevant execution history. 

\vspace{2pt}
\noindent\textbf{Ablation Studies on Self-Supervised Training Objectives.}
As shown in Table~\ref{tab:ss_dtpe_objective_ablation}, adding
Property~\textit{ii)} to Property~\textit{i)} further improves LIBERO-Long
from \(95.0\%\) to \(97.0\%\) and the real-world Long.\ result from \(59\%\)
to \(67\%\). This confirms that monotonic order consistency is particularly
beneficial for maintaining coherent advancement throughout multi-stage
execution. Combining Semantic Progress Alignment with Temporal Structure
Learning achieves the best result in every column. Compared with removing all
objectives, the full formulation improves the four LIBERO suites by \(3.6\),
\(1.4\), \(2.2\), and \(3.4\) percentage points, respectively, and raises the
real-world Long.\ and Gen.\ results by \(18\) and \(12\). These results
demonstrate that semantic alignment and temporal structure provide complementary
supervision.

\begin{table}[t]
    \caption{\textbf{Ablation on HPIM.}
    The upper block evaluates the inter-progress gate and intra-progress
    relevance. \textit{w/o Inter.} fixes \(d_t=1\), applying local relevance
    at all execution stages; \textit{w/o Intra.} replaces token-wise relevance
    with uniform future modulation. The lower block compares a uniform
    layer-wise schedule with our middle-emphasized schedule.}
    \label{tab:hpim_ablation}
    \vspace{-5pt}
    \centering
    \footnotesize
    \setlength{\tabcolsep}{5.5pt}

    \begin{tabular}{l|cccc|cc}
        \toprule
        \multirow{2}{*}{\textbf{Setting}}
        & \multicolumn{4}{c|}{\textbf{LIBERO}}
        & \multicolumn{2}{c}{\textbf{Real-World}} \\
        \cmidrule(lr){2-5}
        \cmidrule(lr){6-7}

        & \textbf{Spatial}
        & \textbf{Object}
        & \textbf{Goal}
        & \textbf{Long}
        & \textbf{Long.}
        & \textbf{Gen.} \\
        \hline

        \rowcolor{gray!20}
        \multicolumn{7}{c}{\textit{Hierarchical Modulation}} \\
        \hline

        w/o Inter.
        & 96.8 & 99.4 & 97.8 & 95.4
        & 58 & 61 \\

        w/o Intra.
        & 97.4 & 99.6 & 97.4 & 95.8
        & 62 & 64 \\
 \rowcolor[HTML]{ECDFF2}
        \textbf{Full HPIM}
    & \textbf{99.6}
        & \textbf{100.0}
        & \textbf{98.8}
        & \textbf{97.8}
        & \textbf{72}
        & \textbf{69} \\
        \hline

        \rowcolor{gray!20}
        \multicolumn{7}{c}{\textit{Layer-Wise Schedule}} \\
        \hline

        Uniform
        &99.2 & 99.6 & 98.2& 97.0
        & 70 & 66 \\

        \rowcolor[HTML]{ECDFF2}
        \textbf{Ours}
        & \textbf{99.6}
        & \textbf{100.0}
        & \textbf{98.8}
        & \textbf{97.8}
        & \textbf{72}
        & \textbf{69} \\
        \bottomrule
    \end{tabular}

    \vspace{-5pt}
\end{table}

\vspace{2pt}
\noindent\textbf{Ablation Studies on HPIM.}
Table~\ref{tab:hpim_ablation} evaluates the hierarchical modulation and
layer-wise schedule in HPIM. In the \textit{w/o Inter.} setting, the
inter-progress gate is removed by fixing \(d_t=1\), while
\textit{w/o Intra.} replaces token-wise future-latent relevance with uniform
modulation. 
Removing either component consistently degrades performance. Compared with the
full HPIM, \textit{w/o Inter.} decreases LIBERO-Long and real-world Long.\ by
\(2.4\) and \(14\) percentage points, while \textit{w/o Intra.} decreases them by \(2.0\)
and \(10\), respectively. These results demonstrate that the global
progress gate determines when future information should be selectively
utilized, while intra-progress relevance  identifies useful 
latents.
For the layer-wise schedule, replacing our schedule with uniform modulation results in consistent degradation. 

\begin{table}[t]
    \caption{\textbf{Ablations on Progress Conditioning in HPIM.}
    \textit{w/o Progress Dynamics} removes
    \(\Delta Z_t^{\mathrm{prog}}\) while retaining
    \(Z_t^{\mathrm{prog}}\) in the global gate.
    \textit{Global Gate w/o Progress} removes both
    \(Z_t^{\mathrm{prog}}\) and \(\Delta Z_t^{\mathrm{prog}}\) from the
    global gate.
    \textit{Local Relevance w/o Progress} removes progress conditioning from
    the local relevance branch. Other components are unchanged.}
    \label{tab:hpim_progress_condition_ablation}
    \vspace{-5pt}
    \centering
    \footnotesize
    \setlength{\tabcolsep}{1.2pt}

    \begin{tabular}{l|cccc|cc}
        \toprule
        \multirow{2}{*}{\textbf{Setting}}
        & \multicolumn{4}{c|}{\textbf{LIBERO}}
        & \multicolumn{2}{c}{\textbf{Real-World}} \\
        \cmidrule(lr){2-5}
        \cmidrule(lr){6-7}

        & \textbf{Spatial}
        & \textbf{Object}
        & \textbf{Goal}
        & \textbf{Long}
        & \textbf{Long.}
        & \textbf{Gen.} \\
        \hline

        w/o Progress Dynamics
        & 98.0 & 98.4 & 97.0 & 96.8
        & 59 & 61 \\

        Global Gate w/o Progress
        & 97.2 & 98.0 & 96.6 & 96.0
        & 57 & 58 \\

        Local Relevance w/o Progress
        & 97.4 & 99.0 & 97.6 & 95.8
        & 63 & 57 \\
        \hline

        \rowcolor[HTML]{ECDFF2}
        \textbf{Full HPIM}
        & \textbf{99.6}
        & \textbf{100.0}
        & \textbf{98.8}
        & \textbf{97.8}
        & \textbf{72}
        & \textbf{69} \\
        \bottomrule
    \end{tabular}

    \vspace{-3pt}
\end{table}
\begin{table}[t]
    \caption{\textbf{Ablation Studies on Attention Modulation Paths.}
    \textit{Action-to-All-Visual} modulates action attention to both current
    and future visual tokens; \textit{All-to-Future} modulates attention from
    all query tokens to future visual tokens; \textit{Action-to-Future} only
    modulates attention from action tokens to future visual tokens, as adopted
    in our ProWAM.}
    \label{tab:modulation_path_ablation}
    \vspace{-5pt}
    \centering
    \footnotesize
    \setlength{\tabcolsep}{3.5pt}

    \begin{tabular}{l|cccc|cc}
        \toprule
        \multirow{2}{*}{\textbf{Modulation Path}}
        & \multicolumn{4}{c|}{\textbf{LIBERO}}
        & \multicolumn{2}{c}{\textbf{Real-World}} \\
        \cmidrule(lr){2-5}
        \cmidrule(lr){6-7}

        & \textbf{Spatial}
        & \textbf{Object}
        & \textbf{Goal}
        & \textbf{Long}
        & \textbf{Long.}
        & \textbf{Gen.} \\
        \hline

        Action-to-All-Visual
        & 98.8 & 99.0 & 97.8 & 96.4
        & 65 & 63 \\

        All-to-Future
        & 98.4 & 98.8 & 97.4 & 96.6
        & 64 & 67 \\
        \hline

        \rowcolor[HTML]{ECDFF2}
        \textbf{Action-to-Future}
        & \textbf{99.6}
        & \textbf{100.0}
        & \textbf{98.8}
        & \textbf{97.8}
        & \textbf{72}
        & \textbf{69} \\
        \bottomrule
    \end{tabular}

    \vspace{-5pt}
\end{table}

\vspace{2pt}
\noindent\textbf{Ablation Studies on Progress Conditioning in HPIM.}
Table~\ref{tab:hpim_progress_condition_ablation} evaluates the role of progress
signals in HPIM. Removing \(\Delta Z_t^{\mathrm{prog}}\) decreases the
real-world Long.\ and Gen.\ results by \(13\) and \(8\) percentage points, indicating that
progress dynamics provide useful cues for execution-stage transitions. Removing
progress information entirely from the global gate leads to larger drops of
\(15\) and \(11\) percentage points, confirming the importance of progress-conditioned
stage-level regulation.
Similarly, removing progress conditioning from local relevance reduces
LIBERO-Long from \(97.8\%\) to \(95.8\%\) and real-world Gen.\ from \(69\%\)
to \(57\%\). 

\vspace{2pt}
\noindent\textbf{Ablation Studies on Attention Modulation Paths.}
Table~\ref{tab:modulation_path_ablation} investigates the effect of different
progress-conditioned attention modulation scopes. Applying modulation from
action queries to all visual tokens (\textit{Action-to-All-Visual}) degrades
performance relative to our design, with reductions of \(1.4\), \(7\), and
\(6\) percentage points on LIBERO-Long, real-world Long., and Gen.,
respectively. Likewise, extending the modulation to all query tokens attending
to future visual latents (\textit{All-to-Future}) leads to corresponding drops.

\vspace{-6pt}
\subsection{More Qualitative and Quantitative Analysis}
\label{sec:more_analysis}

\begin{table}[t]
    \caption{\textbf{Analysis on Progress Source.}
    \textit{Time-Only} uses normalized execution time as the progress signal, while SS-DTPE estimates
    progress from action-observation feedback and recurrent execution history.}
    \label{tab:progress_source_analysis}
    \vspace{-5pt}
    \centering
    \footnotesize
    \setlength{\tabcolsep}{4.2pt}

    \begin{tabular}{l|cccc|cc}
        \toprule
        \multirow{2}{*}{\textbf{Progress Source}}
        & \multicolumn{4}{c|}{\textbf{LIBERO}}
        & \multicolumn{2}{c}{\textbf{Real-World}} \\
        \cmidrule(lr){2-5}
        \cmidrule(lr){6-7}

        & \textbf{Spatial}
        & \textbf{Object}
        & \textbf{Goal}
        & \textbf{Long}
        & \textbf{Long.}
        & \textbf{Gen.} \\
        \hline

        Time-Only
        & 96.8 & 97.8 & 97.0 & 95.2
        & 56 & 59 \\
        \hline

        \rowcolor[HTML]{ECDFF2}
        \textbf{SS-DTPE (Ours)}
        & \textbf{99.6}
        & \textbf{100.0}
        & \textbf{98.8}
        & \textbf{97.8}
        & \textbf{72}
        & \textbf{69} \\
        \bottomrule
    \end{tabular}

    \vspace{-3pt}
\end{table}

\begin{table}[t]
  \caption{\textbf{Analysis on Global-Gate Supervision.}
The upper block evaluates gate supervision, while the lower examines
robustness to perturbations of the weak labels.}
    \label{tab:gate_supervision_analysis}
    \vspace{-5pt}
    \centering
    \footnotesize
    \setlength{\tabcolsep}{2.5pt}

    \begin{tabular}{l|cccc|cc}
        \toprule
        \multirow{2}{*}{\textbf{Setting}}
        & \multicolumn{4}{c|}{\textbf{LIBERO}}
        & \multicolumn{2}{c}{\textbf{Real-World}} \\
        \cmidrule(lr){2-5}
        \cmidrule(lr){6-7}

        & \textbf{Spatial}
        & \textbf{Object}
        & \textbf{Goal}
        & \textbf{Long}
        & \textbf{Long.}
        & \textbf{Gen.} \\
        \hline

        \rowcolor{gray!20}
        \multicolumn{7}{c}{\textit{Effect of Gate Supervision}} \\
        \hline

        w/o Gate Supervision
        & 97.2 & 98.0 & 96.4 & 95.6
        & 59 & 61 \\
        \rowcolor[HTML]{ECDFF2}

        \textbf{w/ Gate Supervision}
        & \textbf{99.6}
        & \textbf{100.0}
        & \textbf{98.8}
        & \textbf{97.8}
        & \textbf{72}
        & \textbf{69} \\
        \hline

        \rowcolor{gray!20}
        \multicolumn{7}{c}{\textit{Robustness to Label Perturbation}} \\
        \hline

        Perturb. (5\%)
        & 99.6 & 99.8 & 98.8 & 97.4
        & 70 & 69 \\

        Perturb. (10\%)
        & 99.2 & 99.8 & 98.4 & 97.2
        & 70 & 67 \\

        Perturb. (20\%)
        & 98.8 & 99.2 & 98.0 & 96.6
        & 68 & 66 \\

        \bottomrule
    \end{tabular}

    \vspace{-8pt}
\end{table}
\begin{figure}[t]
    \centering
    \includegraphics[width=\columnwidth]{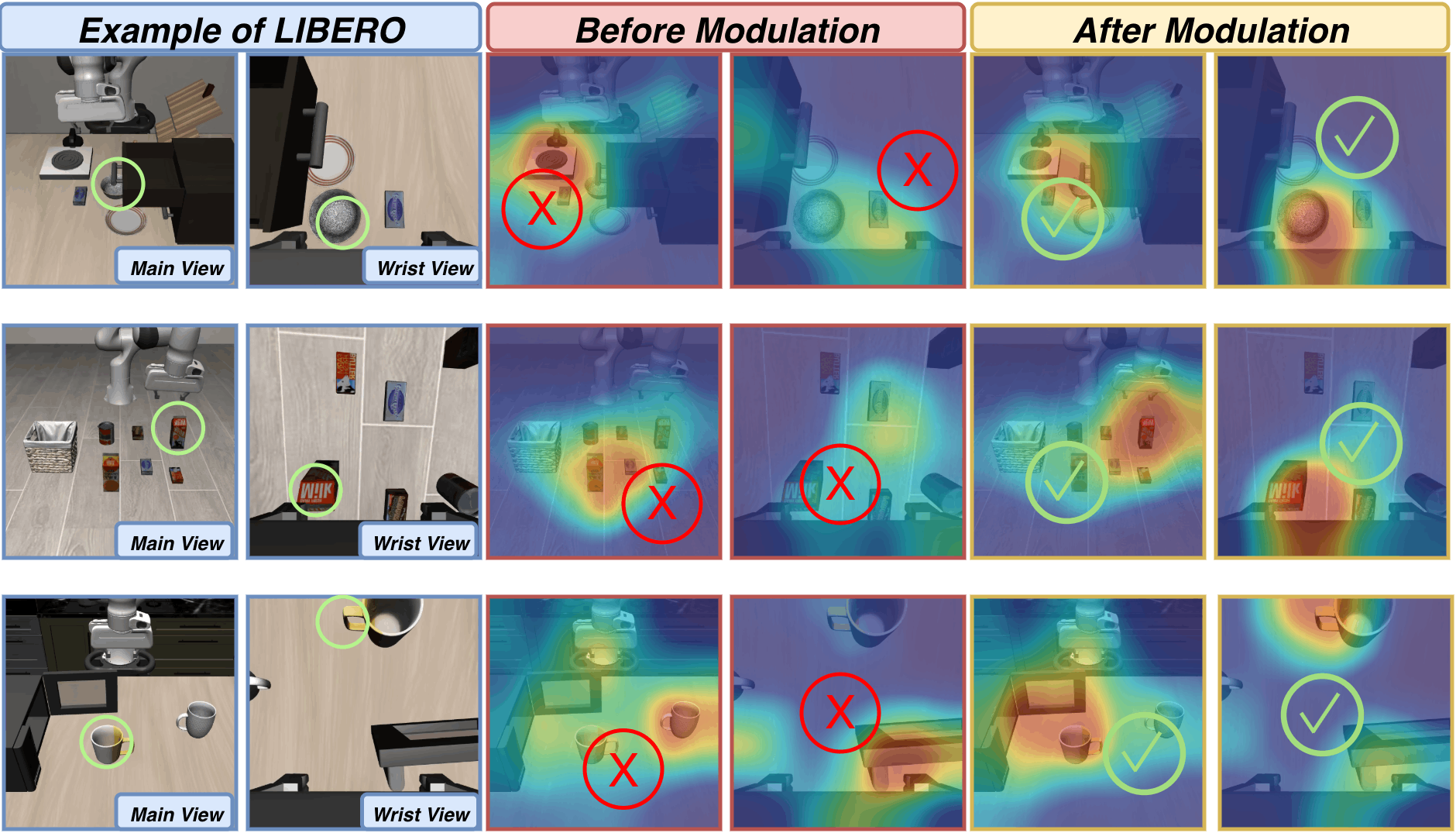}
    \vspace{-13pt}
    \caption{\textbf{Qualitative Visualization of Progress-Conditioned Attention Modulation.}
    Compared with unmodulated attention, HPIM produces more concentrated
responses around instruction-relevant objects and
task-critical interaction regions across both main and wrist views.}
    \label{fig:attention_visualization}
    \vspace{-12pt}
\end{figure}

\begin{figure}[t]
    \centering
       \includegraphics[width=\columnwidth]{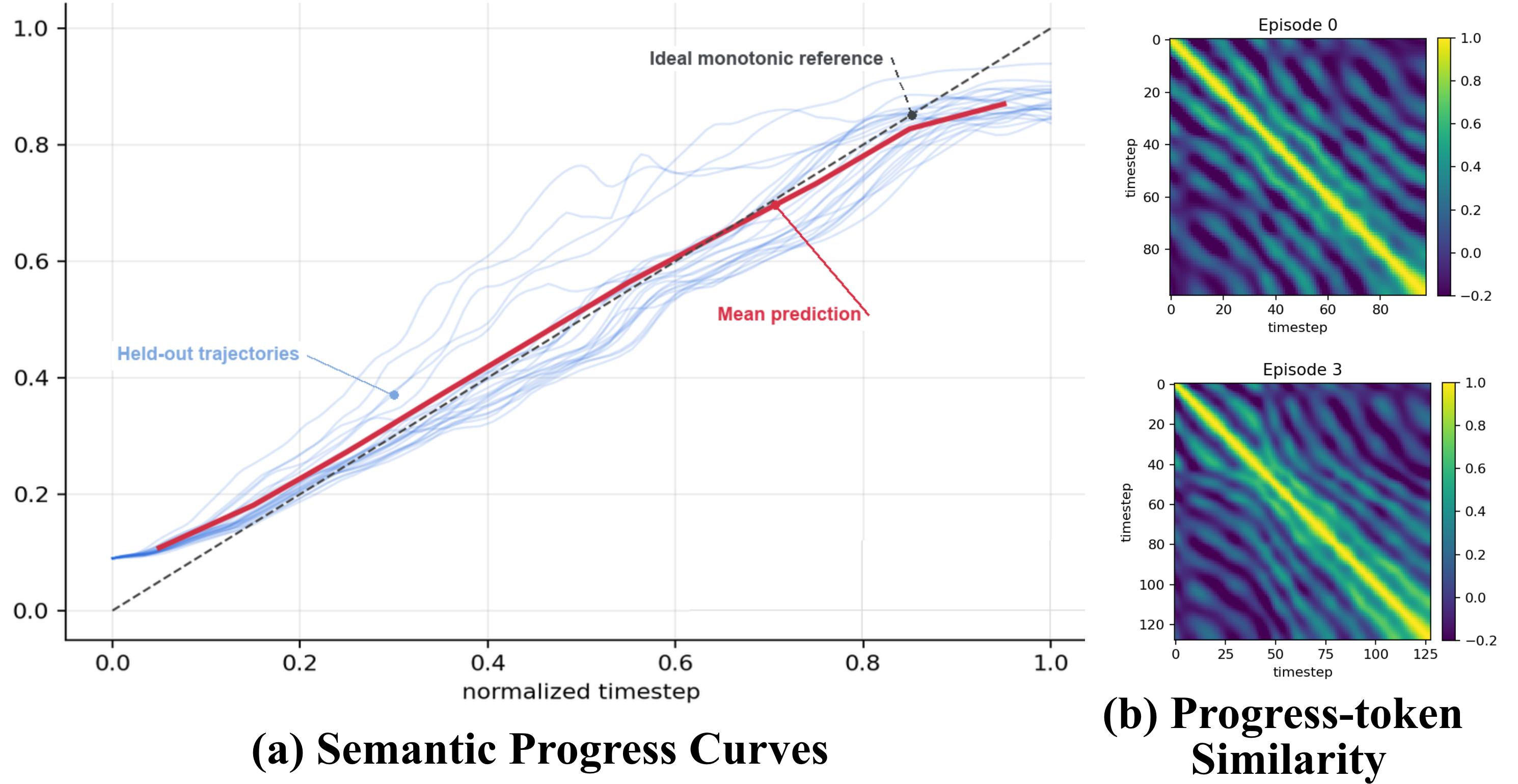}
    \vspace{-15pt}
    \caption{\textbf{Qualitative Visualization of Progress Representations Learned by SS-DTPE.}
    \textbf{(a)} Semantic progress estimates on held-out trajectories exhibit
    consistent and approximately monotonic evolution while preserving
    trajectory-specific variations.
    \textbf{(b)} Progress-token similarity shows strong local temporal coherence
    and lower similarity between temporally distant states.}
    \label{fig:progress_representation_analysis}
    \vspace{-3pt}
\end{figure}

\begin{figure}[t]
    \centering
    \includegraphics[width=\columnwidth]{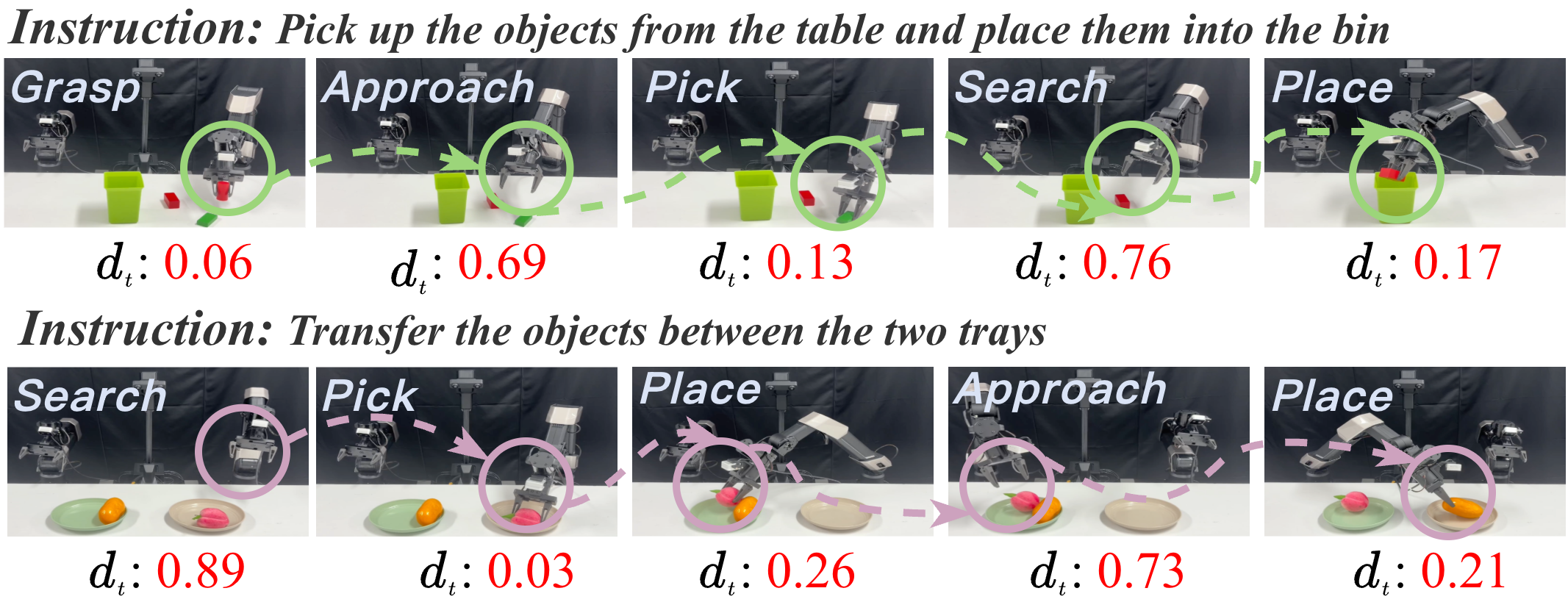}
    \vspace{-15pt}
   \caption{\textbf{Qualitative Visualization of Inter-Progress Global Modulation.}
Representative stages from \textit{Clear Objects into Bin} and
\textit{Transfer Objects from Tray to Tray} show that the global gate \(d_t\)
varies with execution progress, enabling stage-dependent modulation rather than
fixed imagination utilization. Dashed arrows indicate temporal progression.}
    \label{fig:global_gate_visualization}
    \vspace{-5pt}
\end{figure}

\begin{figure*}[!t]
    \centering
    \includegraphics[width=\textwidth]{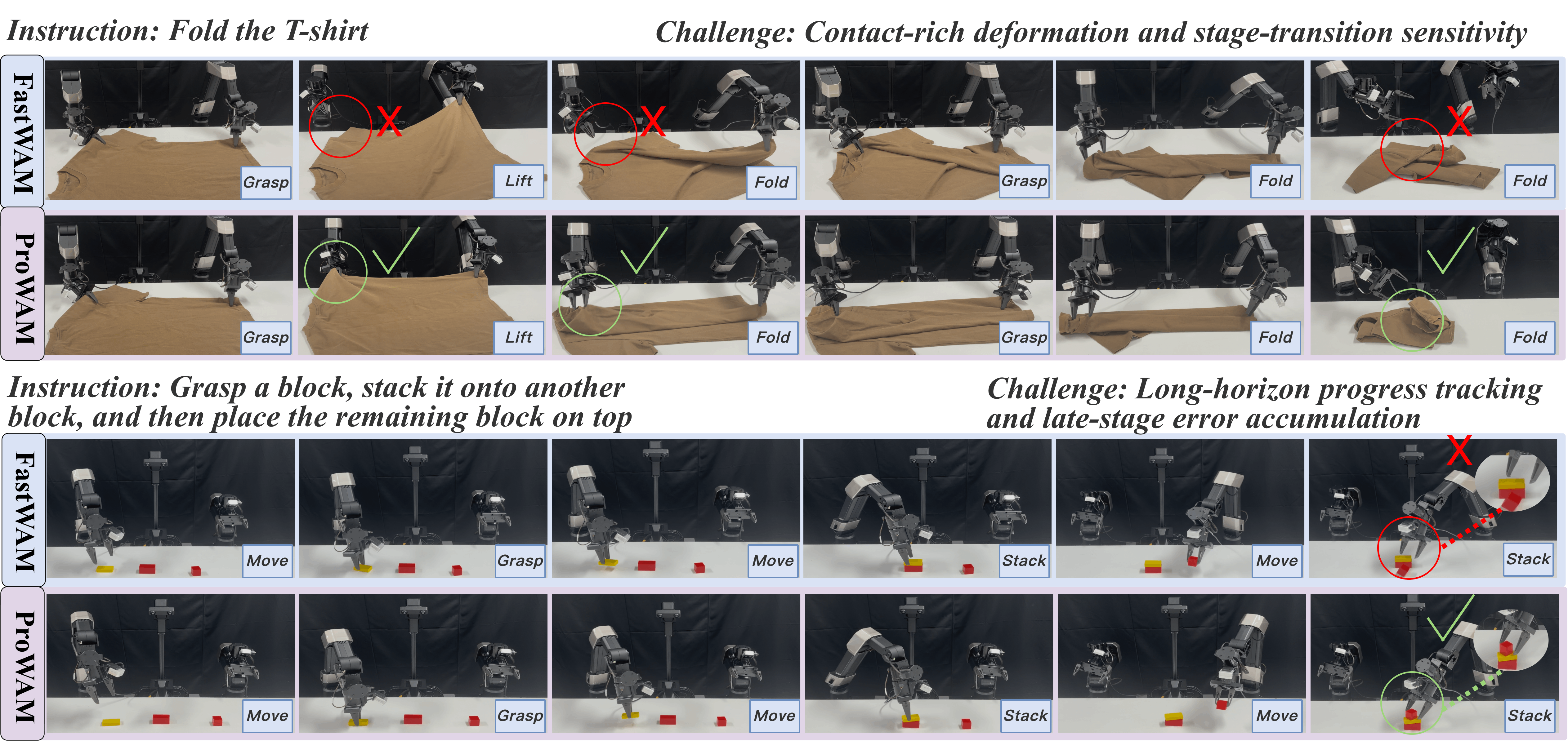}
    \vspace{-15pt}
    \caption{\textbf{Qualitative Comparison on Real-World Manipulation Tasks.}
     Fast-WAM exhibits
    execution errors at critical interaction and transition stages, whereas
    ProWAM maintains more consistent task progression and successfully completes
    the subsequent stages. Red and green annotations highlight 
    failure and successful execution events, respectively.}
    \label{fig:realworld_qualitative}
    \vspace{-10pt}
\end{figure*}

\begin{figure}[t]
    \centering
       \includegraphics[width=\columnwidth]{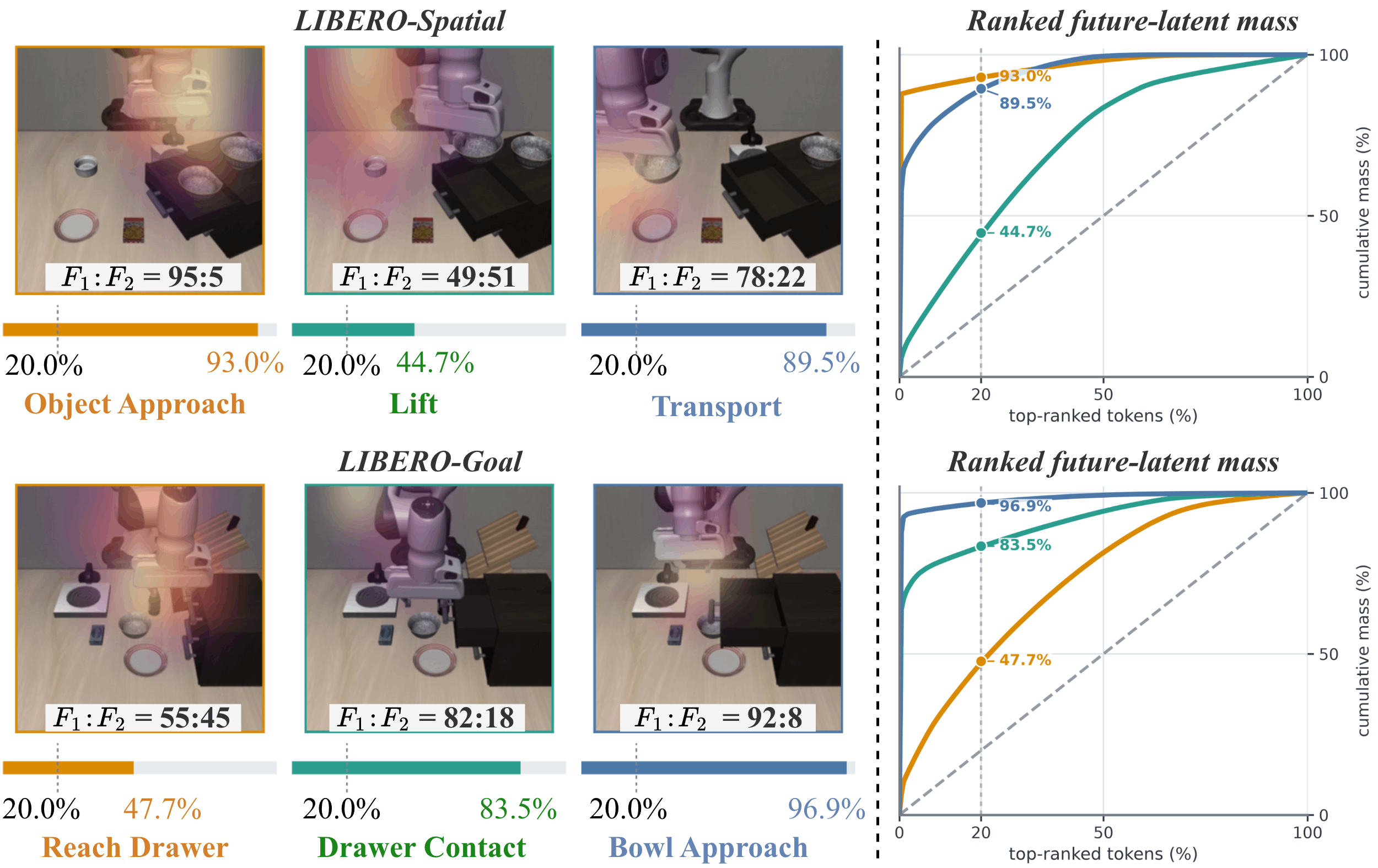}
     \vspace{-13pt}
   \caption{\textbf{Qualitative Visualization of Intra-Progress Future-Latent Relevance.}
\(F_1\!:\!F_2\) denotes the relevance-mass split between the two future frames.
The curves show cumulative relevance over ranked future latents,
with the gray diagonal indicating uniform allocation; marked values report the
mass captured by the top \(20\%\) latents.}
      \label{fig:hpim_future_token_analysis}
    \vspace{-14pt}
\end{figure}


\noindent\textbf{Analysis on Progress Source.}
Table~\ref{tab:progress_source_analysis} compares SS-DTPE with a
\textit{Time-Only} progress signal based solely on normalized execution time.
The latter consistently underperforms SS-DTPE across all evaluation settings.
In particular, SS-DTPE improves LIBERO-Long from \(95.2\%\) to \(97.8\%\),
and achieves gains of \(16\) and \(10\) percentage points on the real-world
Long.\ and Gen.\ tasks, respectively. These results suggest that execution
progress cannot be sufficiently characterized by temporal position alone.
Instead, incorporating action-observation feedback and recurrent execution
history enables SS-DTPE to capture more informative, task-dependent progress
states for downstream modulation.

\vspace{2pt}
\noindent\textbf{Analysis on Global-Gate Supervision.}
Table~\ref{tab:gate_supervision_analysis} evaluates the effectiveness and
robustness of the weak supervision for the global gate. Removing gate
supervision reduces real-world Long.\ and Gen.\ performance by \(13\) and
\(8\) percentage points, respectively. Meanwhile, the model remains robust to label
perturbations, retaining \(68\%\) and \(66\%\) success rates even at \(20\%\)
perturbation. These results demonstrate that the proposed supervision provides
effective stage-level guidance while remaining tolerant to imperfect weak
labels.

\vspace{2pt}
\noindent\textbf{Qualitative Analysis on Progress-Conditioned Attention Modulation.}
Fig.~\ref{fig:attention_visualization} qualitatively illustrates the effect
of HPIM on attention distributions across representative manipulation scenes.
Before modulation, attention is relatively diffuse and may respond strongly to
task-irrelevant regions. In contrast, progress-conditioned modulation produces
more concentrated responses around instruction-relevant objects and
task-critical interaction regions across both main and wrist views. This
consistent refinement across diverse scenes indicates that HPIM effectively
suppresses distracting information while preserving task-relevant visual cues.

\noindent\textbf{Qualitative Analysis on Progress Representation.}
Fig.~\ref{fig:progress_representation_analysis} visualizes the progress
representations learned by SS-DTPE. As shown in
Fig.~\ref{fig:progress_representation_analysis}(a), semantic progress evolves
consistently and approximately monotonically on held-out trajectories, in line
with Semantic Progress Alignment. In
Fig.~\ref{fig:progress_representation_analysis}(b), strong near-diagonal
similarity and reduced similarity between distant states indicate local temporal
coherence and discriminative progress structure.

\vspace{2pt}
\noindent\textbf{Qualitative Analysis on Inter-Progress Global Modulation.}
Fig.~\ref{fig:global_gate_visualization} visualizes the global gate \(d_t\)
along representative real trajectories. Across both tasks, \(d_t\)
adapts consistently with execution progress, assigning higher values during
search and transit-like stages and lower values during contact-rich
interactions. This behavior reflects the changing control demands across
execution stages and demonstrates progress-dependent modulation of imagined
 information.

\vspace{2pt}
\noindent\textbf{Qualitative Analysis of Intra-Progress Future-Latent Relevance.}
Fig.~\ref{fig:hpim_future_token_analysis} visualizes the normalized local
relevance learned by HPIM. For each execution state, we first normalize the
relevance coefficients over all future latents such that their total mass sums
to one. To characterize temporal allocation, we sum the normalized relevance
over all spatial positions and camera views within each future frame, yielding
the \(F_1\!:\!F_2\) relevance split. To measure latent-level concentration, we
then flatten future latents across frames, views, and spatial positions, rank
them by relevance, and accumulate the mass of the top-ranked \(20\%\). Under
uniform allocation, this subset would contain only \(20\%\) of the total mass.
The resulting distributions are highly non-uniform and vary with execution
progress. On LIBERO-Spatial, the top-\(20\%\) mass changes from \(93.0\%\)
during \textit{Object Approach} to \(44.7\%\) during \textit{Lift}, and then to
\(89.5\%\) during \textit{Transport}, accompanied by clear shifts in the
\(F_1\!:\!F_2\) allocation. Similar stage-dependent variations are observed on
LIBERO-Goal. These are consistent with our motivating observation that
imagined future latents exhibit heterogeneous relevance within a progress
state, showing that HPIM adaptively redistributes future-latent relevance as
execution progresses.

\vspace{2pt}
\noindent\textbf{Qualitative Analysis on Real-World Execution.}
Fig.~\ref{fig:realworld_qualitative} compares Fast-WAM and ProWAM on two
representative real-world tasks. In \textit{Stack Three Blocks}, Fast-WAM
accumulates errors toward the final stacking stage, whereas ProWAM maintains
stable progress and completes the full sequence. In \textit{Fold Cloth},
Fast-WAM exhibits unstable behavior during contact-rich lifting and folding,
while ProWAM preserves more coherent execution across successive stages.

\vspace{-10pt}
\section{Conclusion} \label{sec:conclusion} In this work, we propose ProWAM,  a progress-conditioned world action model that
learns when and how to use imagined future information for robotic
manipulation. Motivated by the observation that future visual latents are not
uniformly useful throughout execution, ProWAM treats execution progress as a
principled conditioning signal for adaptive future utilization. Specifically,
SS-DTPE learns structured progress representations by integrating short-term
action--observation feedback with long-term interaction history, while HPIM
hierarchically modulates the action-to-future attention pathway according to
both execution stages and individual future latents. Across five simulation
benchmarks and two real-world platforms, ProWAM demonstrates strong
performance. 

\vspace{-5pt}

\ifCLASSOPTIONcaptionsoff
  \newpage
\fi
\bibliographystyle{IEEEtran}
\bibliography{main} 

\clearpage
\twocolumn[
\begin{center}
    {\Large\bfseries Appendix of ProWAM}
\end{center}
\vspace{1em}
]

\title{Appendix of ProWAM}
\maketitle

\IEEEdisplaynontitleabstractindextext
\IEEEpeerreviewmaketitle

\setcounter{section}{0}
\renewcommand{\thesection}{\arabic{section}}

\section{Detailed Experimental Setup}
\label{sec:supp_exp_setup}

This section provides additional details of the simulation benchmarks,
implementation settings, and evaluation protocols used in the main paper.
\vspace{-7pt}
\subsection{Simulation Benchmarks}
\label{sec:supp_simulation}
LIBERO~\cite{liu2023libero} is a language-conditioned manipulation benchmark built on the Franka
robot platform. We evaluate ProWAM on four suites:
LIBERO-Spatial, LIBERO-Object, LIBERO-Goal, and LIBERO-Long. Each task provides 50 demonstrations, covering general and long-horizon tasks. We report success rates for each suite, evaluated over a total of 2000 trials across 40 tasks with different random seeds.

RoboTwin2.0~\cite{chen2025robotwin}
 is a challenging bimanual manipulation benchmark with over 50 tasks
requiring coordinated dual-arm control. Following the multi-task training setup
of~\cite{li2026causal, bi2026motus}, all models are trained on a mixture of 2,500
demonstrations from clean scenes and 25,000 demonstrations collected under
heavy scene randomization. We report average success rates over 100 trials per task.

VLABench~\cite{zhang2025vlabench} is a large-scale
language-conditioned manipulation benchmark. It contains 100 task categories and over 2,000 objects, covering
semantic instruction following, physical reasoning, and
long-horizon planning. Following the standard protocol, we evaluate ten tasks with 50 episodes per task.
Besides Success Rate (SR), we report Intention Score (IS) and
Progress Score (PS), where PS directly
assesses progress-aware manipulation.

RoboEval~\cite{wang2025roboeval} comprises eight bimanual task families, each
with 3--5 variations covering static settings, position and orientation
perturbations, their combinations, and task-specific configurations. We evaluate
each task--variation pair over 100 episodes. In addition to Success Rate (SR),
we report Task Progression (TP) to measure stage-wise completion and
progress-aware execution.

Mikasa-Robo~\cite{cherepanov2025memory} focuses on temporal robotic manipulation with a Franka robot. It includes 5 memory-dependent tasks, each with 250 demonstrations. We follow the standard protocol~\cite{shi2026memoryvla++} with 250 demonstrations per task, 128 × 128 image observations, and 100 evaluation episodes per task.
\begin{figure}[t]
    \centering
    \includegraphics[width=\columnwidth]{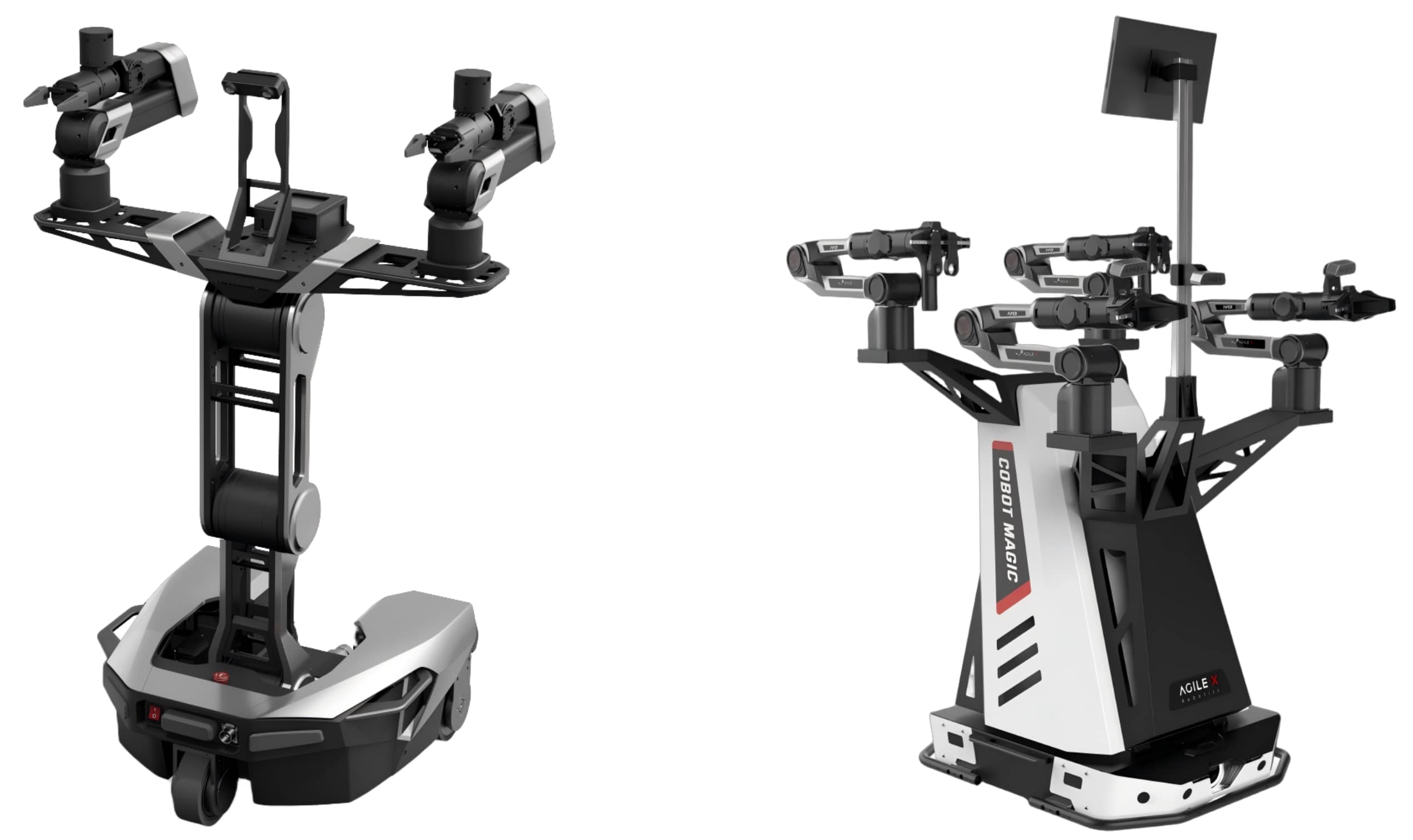}
    \vspace{-8pt}
    \caption{\textbf{Real-world Robotic Platforms Used in Our Experiments.}
    \textbf{Left:} Galaxea R1 Lite.
    \textbf{Right:} AgileX Cobot Magic.
    The two platforms provide complementary mobile-manipulation embodiments for
    evaluating progress-aware long-horizon manipulation and generalization.}
    \label{fig:robot_platforms}
    \vspace{-6pt}
\end{figure}
\subsection{Detailed Real-World Setup}
\label{sec:supp_realworld}

\vspace{2pt}
\noindent\textbf{Robot Platforms.}
As shown in Fig.~\ref{fig:robot_platforms}, we conduct real-world experiments
on two complementary robotic platforms. The \textbf{Galaxea R1 Lite}
platform, developed by Galaxea Dynamics, is a mobile bimanual
manipulation system equipped with dual 6-DoF A1X arms, an omnidirectional
mobile base, and a height-adjustable torso. Its perception system integrates
multiple external and wrist-mounted cameras to provide both global scene
context and interaction-centric observations. The platform supports whole-body
teleoperation and is used for both demonstration collection and closed-loop
policy evaluation.

The \textbf{AgileX Cobot Magic} platform is a Mobile
ALOHA-style mobile manipulation system built on an AgileX Tracer
differential-drive base and equipped with collaborative manipulators and depth
sensors. It supports whole-body teleoperation and coordinated manipulation,
providing a distinct embodiment for evaluating the robustness and
cross-platform generality of ProWAM. Together, the two platforms cover
different mobile bases, manipulation configurations, and sensing setups,
enabling comprehensive evaluation of progress-conditioned manipulation in real-world environments.

\noindent\textbf{Progress-Aware Long-Horizon Tasks.}
We evaluate ProWAM on eight real-world long-horizon manipulation tasks that
require coherent progression across multiple execution stages. These tasks
cover container manipulation, contact-rich deformable-object interaction,
fine-grained insertion, sequential stacking, spatial transfer, and bimanual
handover. Each task is decomposed into progressive execution stages, and
cumulative stage-completion rates are reported to characterize intermediate
task advancement and error accumulation throughout execution. The detailed task
definitions are provided below.

\begin{itemize}

    \item \textbf{Drawer Manipulation}\\
    \textit{Instruction: ``Open the drawer, place the toy into the drawer, and then close
    it.''}\\
    This multi-stage task evaluates sequential container manipulation and
    progress maintenance across distinct interaction phases. The execution is
    divided into three stages:
    1) \textit{``Pull''} -- pulling the drawer open to expose the target region;
    2) \textit{``+Place''} -- placing the target object inside the opened
    drawer; and
    3) \textit{``+Push''} -- closing the drawer after successful placement.
    Successful completion requires the policy to preserve task state across
    opening, object interaction, and closing operations.

    \item \textbf{Fold Cloth (In-distribution)}\\
    \textit{Instruction:``Fold the T-shirt.''} \\
    This deformable object manipulation task examines fabric handling and structured folding execution, demanding comprehensive multi-step planning and sustained task coherence across complex operation chains. Following the established three-stage protocol: 1) \textit{``Grasp''} – firmly grasping the T-shirt edge and executing the first controlled folding motion across the garment surface, 2) \textit{``Fold''} – repeating the folding motion to reinforce the fold, and 3) \textit{``Release''} – grasping the right side of the T-shirt and folding it over to the left side. 

   \item \textbf{Plate Lemon Apple}\\
\textit{Instruction: ``Place both the lemon from the bowl and the apple on the
table onto the plate, then put the lid on the bowl.''}\\
This multi-stage task evaluates long-horizon object transfer and container
interaction across semantically distinct execution stages. The sequence
comprises:
1) \textit{``Open''} -- removing the lid to access the lemon inside the bowl;
2) \textit{``+Retrieve''} -- transferring the lemon from the bowl and the apple
from the table onto the target plate; and
3) \textit{``+Cover''} -- placing the lid back onto the bowl.
Successful execution requires the policy to maintain coherent task progress
across container access, object transfer, and restoration.

   \item \textbf{Object Packing}\\
    \textit{Instruction:``Pack the object on the table into the bag.''} \\
    This dual-arm sequential manipulation task evaluates coordinated bimanual operation for container handling and object packing, requiring precise temporal coordination and spatial reasoning. The progressive stages include: 1) \textit{``Open''} – securely opening the bag and maintaining its accessible configuration, and 2) \textit{``+Pack''} – accurately placing the tabletop item into the bag while maintaining bag stability.

      \item \textbf{Markers Collection}\\
    \textit{Instruction:``Put the markers into the cup.''} \\
    This task assesses the limits of fine-grained precision manipulation through slender object handling and constrained placement, demanding high precision and delicate handling. The two-stage operation involves: 1) \textit{``Insert1''} – picking up the first marker and inserting it into the cup with precise orientation control, and 2) \textit{``+Insert2''} – repeating the process for the second marker while avoiding collisions with the first.

    \item \textbf{Stack Three Blocks}\\
    \textit{Instruction: ``Grasp a block, stack it onto another block, and then
    place the remaining block on top.''}\\
    This sequential stacking task evaluates long-horizon progress tracking and
    late-stage placement accuracy. The execution stages are:
    1) \textit{``Grasp''} -- acquiring the first target block;
    2) \textit{``+Stack I''} -- completing the first stacking operation; and
    3) \textit{``+Stack II''} -- placing the remaining block on top to complete
    the three-block stack.
    Errors accumulated during earlier manipulation stages can directly affect
    the final stacking operation, making persistent progress tracking critical.

    \item \textbf{Transfer Objects from Tray to Tray}\\
    \textit{Instruction: ``Transfer the objects between the two trays.''}\\
    This task evaluates sequential cross-tray object transfer under changing
    spatial configurations. The robot repeatedly performs object acquisition,
    transport, and placement between the two trays. The progressive stages are:
    1) \textit{``Pick''} -- acquiring the target object from its source tray;
    2) \textit{``+Transfer''} -- transporting the object toward the opposite
    tray; and
    3) \textit{``+Place''} -- placing the object at the target location.
    The task requires the policy to update its execution state as source and
    target configurations change during the transfer process.

   \item \textbf{Plate Handover}\\
    \textit{Instruction:``Place the plate on the right into the basket on the left.''} \\
    This dual-arm handover task evaluates dynamic object transfer and inter-arm coordination, focusing on seamless spatial-temporal synchronization during object exchange. The three-stage sequence comprises: 1) \textit{``Pick''} – grasping the target plate securely with the right arm, 2) \textit{``+Pass''} – smoothly transferring the plate to the left arm through a coordinated handover maneuver that maintains object stability throughout the exchange, and 3) \textit{``+Place''} – accurately positioning the plate onto the left-side target basket.

\end{itemize}

\noindent\textbf{Generalization and Robustness Tasks.}
We further evaluate ProWAM on four real-world tasks under controlled variations
in object appearance, geometry, position, lighting, and scene configuration.
These settings assess whether the learned progress representation and
progress-conditioned imagination utilization remain reliable beyond the
nominal training conditions. Detailed task definitions are as below.

\begin{itemize}

    \item \textbf{Sort Blocks by Color}\\
    \textit{Instruction: ``Sort the blocks by color.''}\\
    This task evaluates semantic grounding and spatial generalization under
    variations in object appearance and initial configuration. The execution is
    decomposed into three stages:
    1) \textit{``Pick''} -- locating and grasping the target block;
    2) \textit{``+Sort''} -- transferring the block toward the region
    corresponding to its color; and
    3) \textit{``+Place''} -- placing the block at the correct target location.
    Successful execution requires instruction-consistent manipulation while
    adapting to randomized object positions and scene configurations.

    \item \textbf{Clear Objects into Bin}\\
    \textit{Instruction: ``Pick up the objects from the table and place them
    into the bin.''}\\
    This task evaluates robust multi-object manipulation under changing object
    layouts. The execution consists of:
    1) \textit{``Pick''} -- locating and grasping a target object from the
    workspace; and
    2) \textit{``+Transfer''} -- transporting and placing the object into the
    target bin.
    Because objects are removed sequentially, the scene configuration evolves
    throughout execution, requiring the policy to adapt to both current
    progress state and  remaining objects.

    \item \textbf{Fold Cloth (Out-of-Distribution)}\\
    \textit{Instruction: ``Fold the T-shirt.''}\\
    This out-of-distribution setting evaluates generalization to unseen
    deformable-object properties while preserving the underlying manipulation
    objective. The task follows the same three-stage procedure as the
    in-distribution setting:
    1) \textit{``Grasp''} -- establishing a stable grasp on the designated cloth
    region;
    2) \textit{``+Fold''} -- executing the required folding motion; and
    3) \textit{``+Release''} -- releasing the cloth in the target configuration.
    During evaluation, the T-shirt differs from the training distribution in
    visual appearance and physical dimensions, including unseen colors and
    altered sizes. This setting tests whether coherent execution progress can be
    maintained under appearance and geometry shifts.

    \item \textbf{Stack Bowl on Plate}\\
    \textit{Instruction: ``Stack the bowl on the plate.''}\\
    This task evaluates object-level and spatial generalization under variations
    in object configuration and scene conditions. The execution consists of:
    1) \textit{``Pick''} -- locating and securely grasping the target bowl; and
    2) \textit{``+Place''} -- positioning the bowl onto the target plate.
    Successful execution requires robust object grounding and precise placement
    despite variations in the initial scene configuration.

\end{itemize}


\subsection{Evaluation Metrics}
\label{sec:supp_metrics}

We follow the standard evaluation metrics defined by each benchmark. For
completeness, we summarize the principal metrics used throughout our
experiments.

\noindent\textbf{Success Rate (SR).}
Success Rate measures the fraction of evaluation episodes in which the complete
task is successfully accomplished. Given \(N\) evaluation episodes, we compute
\begin{equation}
    \mathrm{SR}
    =
    \frac{1}{N}
    \sum_{n=1}^{N}
    \mathbb{I}\!\left(s_n=1\right),
\end{equation}
where \(s_n\) denotes the binary success indicator of the \(n\)-th episode and
\(\mathbb{I}[\cdot]\) is the indicator function. For benchmarks containing
multiple tasks or suites, the overall performance is reported as
\begin{equation}
    \mathrm{Avg.}
    =
    \frac{1}{M}
    \sum_{m=1}^{M}
    \mathrm{SR}_m,
\end{equation}
where \(M\) denotes the number of evaluated tasks or suites.

\noindent\textbf{Intention Score (IS) and Progress Score (PS).}
For VLABench, we additionally report Intention Score (IS) and Progress Score
(PS) following the official evaluation protocol. Let
\(\phi_{\mathrm{IS}}(\tau_n)\) and \(\phi_{\mathrm{PS}}(\tau_n)\) denote the
benchmark-defined intention and progress scores for trajectory \(\tau_n\).
Their dataset-level values are summarized as
\begin{equation}
    \mathrm{IS}
    =
    \frac{1}{N}\sum_{n=1}^{N}\phi_{\mathrm{IS}}(\tau_n),
    \qquad
    \mathrm{PS}
    =
    \frac{1}{N}\sum_{n=1}^{N}\phi_{\mathrm{PS}}(\tau_n).
\end{equation}
IS evaluates whether the executed behavior is consistent with the intended
task semantics, whereas PS characterizes intermediate task advancement beyond
terminal success alone. PS therefore provides a particularly informative
measure for evaluating progress-aware manipulation.

\noindent\textbf{Task Progression (TP).}
For RoboEval, we report Task Progression (TP) in addition to Success Rate,
following the official benchmark implementation. RoboEval decomposes each task
into semantically defined execution stages and records stage-wise completion,
thereby providing a fine-grained measure of partial task advancement. Compared
with terminal SR, TP more directly reflects whether a policy can maintain
coherent progression throughout multi-stage execution.

\noindent\textbf{Real-World Stage Completion.}
For real-world long-horizon tasks, we report cumulative stage-completion rates.
Given a task decomposed into \(K\) sequential stages, the completion rate of
stage \(k\) is defined as
\begin{equation}
    S_k
    =
    \frac{1}{N}
    \sum_{n=1}^{N}
    \mathbb{I}
    \left[
        \bigcap_{j=1}^{k}
        \mathcal{S}_{n,j}
    \right],
\end{equation}
where \(\mathcal{S}_{n,j}\) denotes successful completion of stage \(j\) in the
\(n\)-th rollout. Thus, completion of a later stage requires all preceding
stages to be successfully executed. The final-stage rate \(S_K\) corresponds
to the full-task success rate. This cumulative formulation explicitly captures
performance degradation and error accumulation as task execution progresses.

\begin{figure*}[t]
    \centering
    \includegraphics[width=\textwidth]{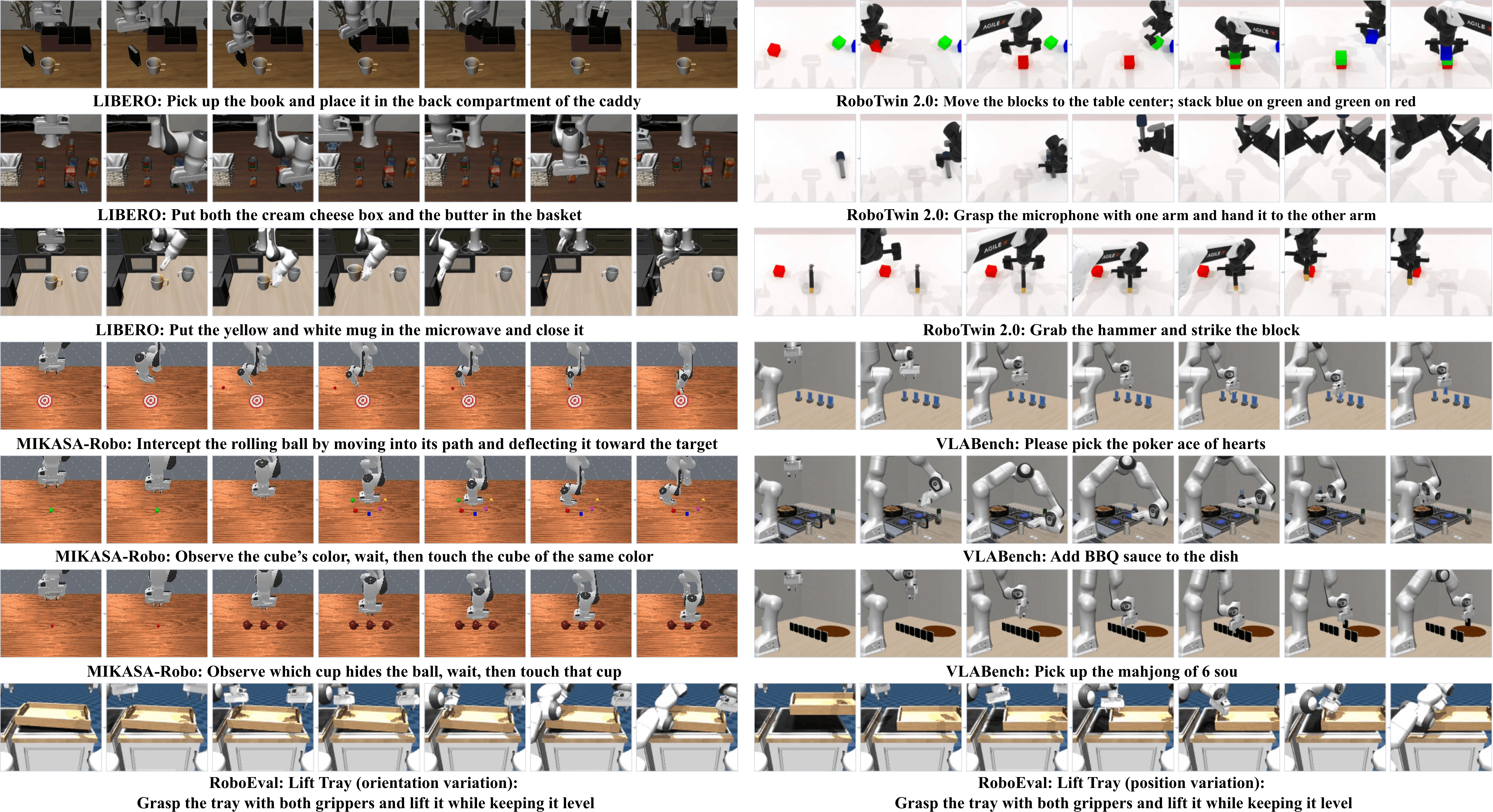}
    \vspace{-15pt}
    \caption{\textbf{Qualitative Simulation Results Across Five Benchmarks.}
    Representative execution sequences of ProWAM on LIBERO, RoboTwin~2.0,
    Mikasa-Robo, VLABench, and RoboEval are shown in temporal order from left
    to right. The examples cover long-horizon object manipulation, bimanual
    coordination and handover, contact-rich interaction, memory-dependent
    control, fine-grained semantic grounding, and robustness to task
    variations.}
    \label{fig:simulation_qualitative}
    \vspace{-5pt}
\end{figure*}

\begin{figure*}[t]
    \centering
    \includegraphics[width=\textwidth]{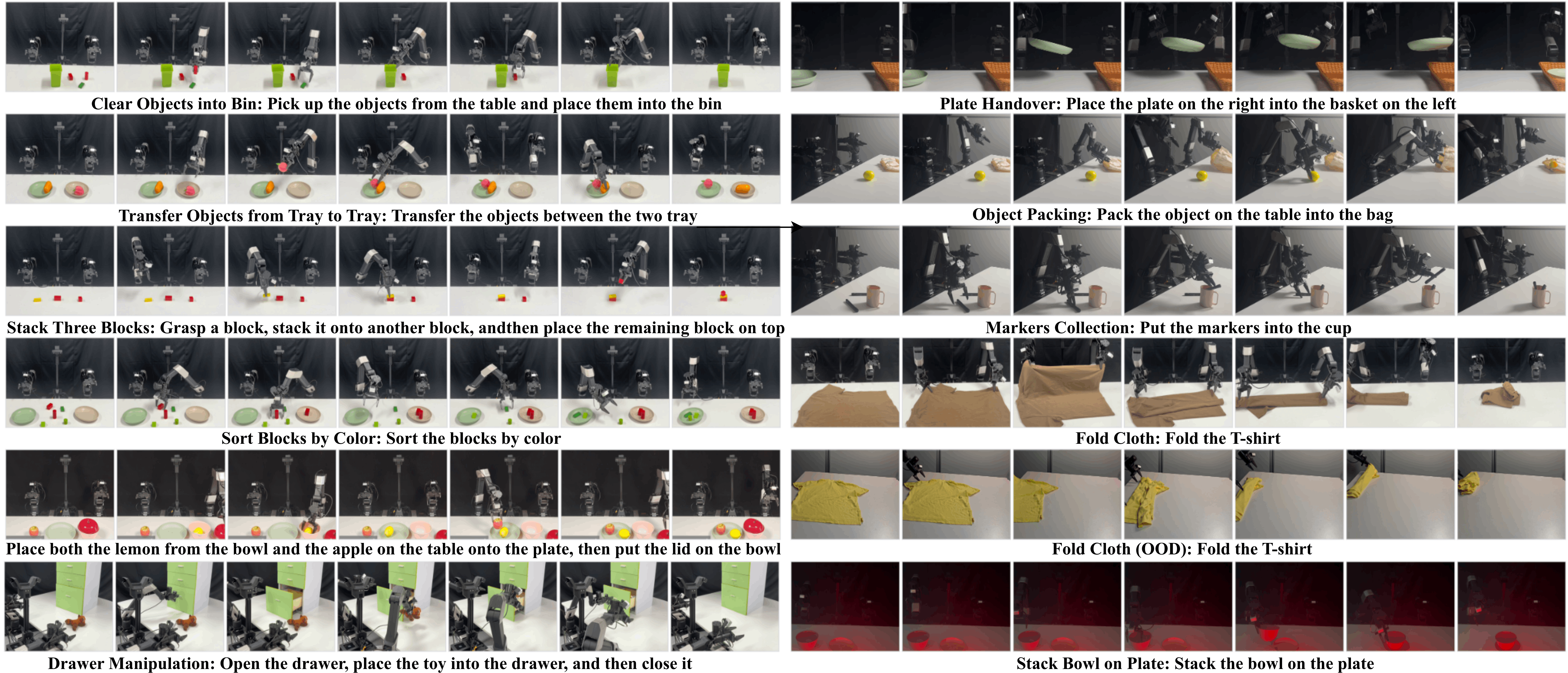}
    \vspace{-15pt}
    \caption{\textbf{Qualitative Results on Real-World Manipulation Tasks.}
    Representative execution sequences of ProWAM across eight progress-aware
    long-horizon tasks and four generalization and robustness tasks are shown
    in temporal order from left to right. The examples cover sequential object
    manipulation, bimanual coordination and handover, deformable-object
    interaction, multi-object transfer, fine-grained insertion, and
    out-of-distribution appearance and scene variations.}
    \label{fig:realworld_qualitative_all}
    \vspace{-5pt}
\end{figure*}

\subsection{Automatic Weak-Supervision Construction for Global Modulation}
\label{sec:supp_global_supervision}

The inter-progress global modulation requires coarse supervision that reflects
the changing control demands across task execution. Rather than manually
annotating progress boundaries for individual trajectories, we automatically
construct weak stage-level supervision from proprioceptive signals available in
robot demonstrations. The key intuition is that contact-rich manipulation can
be approximately localized by characteristic interaction events, whereas the
remaining portions of a trajectory primarily correspond to search, approach,
and transit stages.

\noindent\textbf{Interaction-Anchor Detection.}
For each demonstration trajectory, we first identify interaction anchors from
low-dimensional robot states and actions. In particular, gripper open--close
transitions provide reliable cues for grasping and releasing events. When
gripper transitions alone are insufficient, we additionally exploit
task-dependent end-effector height constraints to identify interaction phases
such as insertion, placement, pouring, or other contact-rich operations. An
optional motion cue based on end-effector velocity is used only when reliable
gripper signals are unavailable. Denoting the corresponding event indicators
at time \(t\) as \(e_t^{g}\), \(e_t^{h}\), and \(e_t^{m}\), respectively, the
interaction anchor is defined as
\begin{equation}
    e_t
    =
    e_t^{g}
    \lor
    e_t^{h}
    \lor
    e_t^{m},
\end{equation}
where \(\lor\) denotes the logical OR operation. For bimanual trajectories,
events detected from either arm are merged to obtain a single trajectory-level
interaction indicator.

\noindent\textbf{Temporal Weak-Label Assignment.}
A single interaction event typically corresponds to a short manipulation phase
rather than an isolated frame. We therefore temporally expand each detected
anchor over a local neighborhood. Specifically, for an interaction event
detected at time \(t\), frames within
\([t-\Delta_{\mathrm{pre}},\,t+\Delta_{\mathrm{post}}]\) are assigned to the
interaction region:
\begin{equation}
    c_{\tau}
    =
    \mathbb{I}
    \left[
        \exists t:
        e_t=1,\;
        \tau\in
        [t-\Delta_{\mathrm{pre}},
         t+\Delta_{\mathrm{post}}]
    \right],
\end{equation}
where \(c_{\tau}=1\) denotes a contact-rich interaction stage and
\(c_{\tau}=0\) denotes a search-, approach-, or transit-dominated stage. We use
\(\Delta_{\mathrm{pre}}=\Delta_{\mathrm{post}}=20\) frames by default. This
temporal expansion provides coarse stage supervision while avoiding subjective
frame-wise annotation of precise interaction boundaries.

Since the global modulation in ProWAM assigns stronger future utilization to
search- and transit-like stages and weaker utilization to contact-rich
interaction stages, we convert the interaction indicator into the global-gate
 as
\begin{equation}
    y_t^{g}=1-c_t,
\end{equation}
such that larger targets correspond to stages in which imagined future
information is expected to be more useful. The resulting weak target is used to
supervise the inter-progress global gate during training.

\vspace{2pt}
\noindent\textbf{Label Refinement and Validity Filtering.}
To reduce noise introduced by instantaneous state fluctuations, we remove
short isolated interaction segments and merge small temporal gaps between
adjacent interaction regions. We further discard unreliable labels when the
required proprioceptive signals are missing or when detected interaction
patterns are inconsistent with the corresponding task configuration.
Consequently, the proposed procedure provides scalable stage-level supervision
directly from demonstration trajectories without requiring manual progress
annotations.

\section{Qualitative Rollout Visualization}

\noindent\textbf{Qualitative Simulation Results.}
Fig.~\ref{fig:simulation_qualitative} presents representative execution
sequences across all five simulation benchmarks, covering a broad spectrum of
manipulation challenges and control regimes. The LIBERO examples involve
multi-stage object relocation, sequential object handling, and container
interaction, where ProWAM maintains coherent execution as task requirements
evolve across stages. RoboTwin~2.0 further stresses coordinated manipulation
through block stacking, bimanual handover, and contact-rich tool use, requiring
the policy to adapt between free-space motion and precise interaction. On
Mikasa-Robo, ProWAM successfully handles both dynamic interception and
memory-dependent tasks, where task-relevant information must be preserved over
extended temporal intervals before the final action is executed. The VLABench
examples illustrate fine-grained language grounding and object discrimination
in visually similar scenes, while RoboEval evaluates stable bimanual
manipulation under changes in tray orientation and position. Across these
settings, the policy exhibits consistent progression from coarse goal-directed
motion to fine-grained interaction, without obvious loss of execution
coherence at stage transitions. These qualitative results provide further
evidence that ProWAM can adapt its use of imagined future information to
different execution demands, consistent with the proposed
progress-conditioned imagination utilization.

\noindent\textbf{Qualitative Real-World Results.}
Fig.~\ref{fig:realworld_qualitative_all} presents representative execution
sequences across all real-world tasks. The long-horizon examples cover
multi-stage manipulation such as object packing, marker insertion, block
stacking, tray transfer, drawer manipulation, and bimanual handover. Across
these tasks, ProWAM maintains coherent execution as the control objective
changes from free-space motion to precise object interaction, and successfully
progresses through successive manipulation stages. Contact-rich tasks such as
cloth folding and handover further require continuous adaptation to interaction
feedback and substantial changes in scene configuration, providing challenging
settings for progress-aware control.

We additionally visualize representative executions under the generalization
and robustness settings. ProWAM successfully performs color-conditioned
sorting, multi-object clearing, out-of-distribution cloth folding, and
bowl-on-plate stacking despite variations in object appearance, geometry, and
scene conditions. In particular, the OOD cloth-folding example introduces
substantial visual and object-level shifts while preserving the underlying
manipulation objective. Overall, these qualitative results indicate that ProWAM
maintains coherent task progression across diverse execution stages and
generalizes its progress-conditioned control behavior beyond the nominal
training conditions.

\section{Limitations and Future Work}
\label{app:limitations_future}

\noindent\textbf{Scope of real-world evaluation.}
Our real-world experiments cover diverse long-horizon manipulation tasks on
two robotic platforms, including sequential object manipulation, bimanual
coordination, deformable-object interaction, and out-of-distribution
generalization. While these settings provide complementary evaluation of
progress-aware manipulation, the current study primarily focuses on
tabletop-scale manipulation scenarios with relatively structured task
objectives.

\noindent\textbf{Future Work 1: Broader embodied task settings.}
An important direction is to extend ProWAM to more diverse embodied settings,
including mobile manipulation, articulated environments, tool-mediated
interaction, and tasks involving longer spatial and temporal dependencies.
Such evaluations would provide a broader characterization of how
progress-conditioned imagination behaves when task progress depends jointly on
navigation, interaction, and manipulation over extended horizons.

\noindent\textbf{Cross-embodiment evaluation scale.}
ProWAM is evaluated on heterogeneous simulated embodiments and two real-world
robotic platforms, demonstrating that the proposed progress-aware formulation
is not tied to a single robot configuration. Nevertheless, large-scale
cross-embodiment transfer is beyond the scope of the present study, as the
training data, observation spaces, and action representations still follow the
configuration of each target benchmark or robotic platform.

\noindent\textbf{Future Work 2: Large-scale cross-embodiment transfer.}
Future work could investigate shared progress representations across robots
with different kinematics, sensing configurations, and action spaces. Since
execution progress describes task advancement at a level complementary to
low-level embodiment-specific control, it provides a promising interface for
studying transferable world-action models. Combining ProWAM with
cross-embodiment pretraining or embodiment-aware action representations may
further improve transfer across robotic platforms with limited target-domain
demonstrations.

\end{document}